\documentclass[10pt]{spieman}  
\usepackage{amsmath,ams fonts,amssymb}
\usepackage{graphicx}
\usepackage{setspace}
\usepackage{tocloft}
\usepackage{algorithmic}
\usepackage{algorithm}
\usepackage{tabularx}
\usepackage{subfig}
\usepackage{multirow}

\title{Automated Resolution Selection for Image Segmentation}

\author[a]{Fares Al-Qunaieer}
\author[b,*]{Hamid R. Tizhoosh}
\author[c]{Shahryar Rahnamayan}
\affil[a]{The National Center for Computer Technology and Applied Mathematics, King Abdulaziz City for Science and Technology (KACST), P.O. Box 6086, Riyadh 11442, Saudi Arabia}
\affil[b]{KIMIA Lab, University of Waterloo, 200 University Ave. West, Waterloo, ON N2L 3G1, Canada}
\affil[c]{The Department of Electrical, Computer, and Software Engineering, University of Ontario Institute of Technology (UOIT), 2000 Simcoe Street North, Oshawa, ON L1H 7K4, Canada}

\begin{document}
\maketitle

\begin{abstract}
It is well-known in image processing in general, and hence in image segmentation in particular, that computational cost increases rapidly with the number and dimensions of the images to be processed. Several fields, such as astronomy, remote sensing, and medical imaging, routinely use numerous very large images, which might also be 3D and/or captured at several frequency bands, all adding to the computational expense. Multiresolution analysis is a method of increasing the efficiency of the segmentation process. One multiresolution approach is the coarse-to-fine segmentation strategy, whereby the segmentation starts at a coarse resolution and is then fine-tuned during subsequent steps. The starting resolution for segmentation is generally selected arbitrarily with no clear selection criteria. The research reported in this paper showed that starting from different resolutions for image segmentation results in different accuracies and computational times, even for images of the same category (depicting similar scenes or objects). An automated method for resolution selection for an input image would thus be beneficial. This paper introduces a framework for the automated selection of the best resolution for image segmentation. We propose a measure for defining the best resolution based on user/system criteria, offering a trade-off between accuracy and computation time. A learning approach is then introduced for the selection of the resolution, whereby extracted image features are mapped to the previously determined best resolution. In the learning process, class (i.e., resolution) distribution is generally imbalanced, making effective learning from the data difficult. A variant of AdaBoost, called RAMOBoost, is therefore used in this research. Local Binary Patterns (LBP) features are utilized in this research. Experiments conducted with three datasets using two different segmentation algorithms show that the resolutions selected through learning enable much faster segmentation than the original ones, while retaining at least the original accuracy. For two of the three datasets used, the segmentation results obtained with the proposed framework were significantly better than with the original resolution with respect to both accuracy and time.
\end{abstract}


{\noindent \footnotesize\textbf{*}Hamid Tizhoosh,  \linkable{tizhoosh@uwaterloo.ca} }

\begin{spacing}{1}   

\section{Introduction}
Routinely processing large images is computationally expensive, specially under real-time constraints. The computation dramatically increases for 3D images and/or in multiple frequency bands. Recently, new technologies enable the capturing of images in much greater detail, which results in big data. Medical imaging is an example in which an imaging examination could require Gigabytes or Terabytes of memory \cite{MedicalLargeImg}. These images can be very large and/or consist of thousands of slices \cite{MedicalLargeImg}. Another example is hyperspectral images capturing hundreds of spectral bands \cite{Hyperspectral}. Furthermore, these images could be processed in batches. One of the challenges for such images is the slow processing time \cite{MedicalLargeImg}. Even a slight increase in speed for processing an image or slice in a set (for 3D or multi-spectral images) can result in a greater overall speed.

In many real-world cases, it may not desirable or affordable to employ powerful hardware for fast processing of big image data. Therefore, several research projects have been conducted on developing software solutions, such as parallel algorithms \cite{Parallel} and multiresolution schemes \cite{WLS-Leroy, multi-Al-Qunaieer, GraphCut_MR}. The multiresolution coarse-to-fine image segmentation has proved to be an efficient method for increasing both the speed and accuracy of  segmentation algorithms as discussed in Section \ref{sec:MultiResImgSeg}.

Some researchers have used image segmentation algorithms in a coarse-to-fine fashion with arbitrarily chosen initial resolutions for the processing of all images \cite{WLS-Tsang, arbitSubsample1, arbitSubsample2, ASM_MR}. However, segmenting an image at a variety of resolutions results in different accuracies and running times, as discussed in \cite{SegDiffRes1} and \cite{SegDiffRes2}. Segmentation results in different accuracies and times per resolution, even with images of the same dataset containing similar scenes/objects as shown in Section \ref{sec:propMeasurePerf}. To the best of the author's knowledge, no work has been reported for automated selection of resolution for image segmentation. Studies have been conducted on scale selection for the scale-space representation, which are reviewed in Section \ref{sec:ScaleSelection}, but they are not directly related to this research, as will be explained.

Defining the best resolution for image segmentation is empirical and directly related to the nature of the application. Two factors can be used for defining the best resolution for image segmentation: accuracy and time. This research was concerned with the simultaneous consideration of both time and accuracy; hence investigating an intelligent approach toward establishing a trade-off between time and accuracy becomes crucial. For critical systems, such as medical applications, accuracy is the primary concern, while in other applications, such as for robot navigation or image retrieval, speed can be more important than high accuracy. There must, therefore, be a trade-off between accuracy and speed. Without an appropriate criterion, of course, selecting a resolution that yields the desired accuracy and speed is difficult.

In this paper, we first propose a measure for selecting the best resolution with respect to the user/system preference. The measure allows to quantify a trade-off between accuracy and time. We show that the resolutions obtained by this measure cannot be fixed, even for images depicting similar scenes/objects. Then, we formulate the resolution selection as a classification problem, where features extracted from an image are mapped to the best resolution determined by the measure. Local Binary Patterns (LBP) are used for the learning of the best resolution, and RAMOBoost, a boosting learning method, has been utilized. Please note that while this work intends to find the best resolution for image segmentation, it is not concerned with developing or modifying image segmentation algorithms, nor with multiresolution analysis methods.

The remaining of this paper is organized as follows: In Section \ref{sec:RelatedWork}, related works on multiresolution image segmentation and scale selection are reviewed. Then, the proposed framework for automated resolution selection is presented in Section \ref{sec:Framework}. Experimental verification of the approach is conducted in Section \ref{sec:Experiments}. Finally, Section \ref{sec:Conclusion} concludes the paper.

\section{Related Work} \label{sec:RelatedWork}
\subsection{Multiresolution Image Segmentation} \label{sec:MultiResImgSeg}
Multiresolution analysis has been extensively used for developing various image segmentation algorithms \cite{multiResSeg1, multiResSeg2, multiResSeg3, multiResSeg4,Qunaieer2011,Qunaieer2014}. Majority of these methods utilize information from some or all resolutions. Although such approach could result in high accuracy, it might also be computationally expensive. In this work, we are interested in a technique which increases image segmentation efficiency, namely, coarse-to-fine segmentation strategy. In this technique, the segmentation process starts from a coarse resolution, and then is tuned at finer ones. Many segmentation algorithms could benefit from this method, because it simplifies the input image rather than modifying the segmentation algorithm. Numerous works in the literature used this method to increase the efficiency of several segmentation algorithms. However, the starting resolution for segmentation is always arbitrarily selected with no apparent generally reusable criteria.

Snakes active contour segmentation algorithms have been used with multiresolution in several research studies. Leroy et al. \cite{WLS-Leroy} enhanced snakes algorithm by using the pyramid representation. The propagation of the curve starts from the coarsest resolution and iteratively continues toward finer resolutions when the curve evolution has converged at previous ones. This method is faster than the original snakes method because most of the calculations are performed at coarse resolutions that requires less computation. Yan et al. \cite{WLS-Yan} proposed an algorithm based on snakes for prostate segmentation in transrectal ultrasound (TRUS) images. They used prior shape models and propagated the curve using Laplacian pyramid in a manner similar to that employed by Leroy et al. to increases the capturing range of the curve and enhance the efficiency of the initialization. Similar approaches were used by Dehmeshki et al. \cite{WLS-Dehmeshki} and Yoon et al. \cite{WLS-Yoon}.

Level set is a well-known active contour approach, but it is computationally expensive and thus slow to converge. Numerous research works have been conducted to solve this problem through the multiresolution coarse-to-fine strategy. Gaussian pyramid was utilized by Tsang \cite{WLS-Tsang} to enhance edge-based level set active contours. The curve is initialized at the coarsest resolution, and propagation proceeds with finer resolutions. A similar methodology has been employed for segmenting ultrasound echocardiographic images \cite{MRLS-Lin1, MRLS-Lin2}. Al-Qunaieer et al. proposed a method for accelerating region-based level set image segmentation \cite{multi-Al-Qunaieer}. The authors used wavelets to decompose the image into three resolutions, with the curve evolution beginning from the coarsest resolution. The results confirmed that using multiresolution approach reduces the effect of noise for large objects and accelerates the convergence rate of the segmentation process.

Graph-based segmentation algorithms have been combined with multiresolution as well to reduce computation time. Roullier et al. \cite{Graph_MR} applied multiresolution approach to graph-based segmentation for mitosis extraction in breast cancer histological whole-slide images. The segmentation begins at a coarse resolution, and at each finer resolution, the resulting segmentation is refined through semi-supervised clustering. Lombaert et al. \cite{GraphCut_MR} adopted a similar approach, but rather than using clustering for the fine-tuning, they applied graph cuts on a narrow banded graph obtained from the resulting minimum cut at the coarser resolution. They showed that their method dramatically increases speed and reduces memory usage without affecting the accuracy of the graph cuts segmentation.

Coarse-to-fine strategy has been incorporated to enhance other segmentation methods, such as MRF segmentation \cite{MRF_MR}, Active Shape Model (ASM) \cite{ASM_MR}, and Active Region algorithm \cite{ARG_MR}.

These studies all offer either a vague or no explanation of the method for selecting the initial resolution to start the segmentation process. Some adopted a trial-and-error technique to select a specific resolution level for all images of a single or different dataset(s), which is impractical because, as shown in Section \ref{sec:propMeasurePerf}, with respect to time and accuracy, the corresponding performance varies over different resolutions, even for images from the same dataset.

\subsection{Scale Selection} \label{sec:ScaleSelection}
The problem of selecting the best scale for image processing has attracted the attention of several researchers. A number of proposed methods described in the literature were based on local information obtained from the derivatives of the image \cite{ScaleSpaceBook, Related-Lindeberg4, Related-Jeong, Related-Elder}. Concepts borrowed from information theory have also been used for scale selection \cite{Related-Jagersand, Related-Sporring, Related-Hadjidemetriou, Related-Kadir}. In addition, statistical methods have been utilized for the selection of the most favourable scales \cite{Related-Marimont, Related-Pedersen, PE}.

In all of these approaches, the objective of scale selection was related primarily to feature detection, with some works also investigating primitive segmentation tasks (e.g., edge detection \cite{Related-Lindeberg4, Related-Jeong, Related-Elder}). Several studies have incorporated scale selection into the image segmentation process. With some modifications, Bayram et al. \cite{ASS} applied the ``minimum reliable scale'' method proposed in \cite{Related-Elder} in order to find the edges in medical images. Lendeberg's scale selection method \cite{Related-Lindeberg4} was utilized by Piovano and Papadopoulo \cite{LS} to guide snakes active contour inside homogeneous regions. Li et al. \cite{SSS} proposed a scale selection method for supervised image segmentation. For each scale in a training image, features are extracted per pixel and assigned to their corresponding labels. Therefore, for $n$ scales, there are $n$ learned classifiers. For a test image, pixels at each scale are classified based on their corresponding classifier, and the best scale is the one at which the posterior probability calculated from the classification is the highest. Although these methods incorporated scale selection in the segmentation process, several distinctions exists between them and the work presented in this paper:

\begin{itemize}
  \item The previous methods were intended for selecting the best Gaussian scale for the scale-space approach. Since scale-space does not involve the subsampling of the image, the grid size (number of pixels) is the same for all scales. For the research for this paper, pyramid scheme is used, in which each resolution is filtered and subsampled from the previous resolution. The reduced size is the major contributing factor in decreasing the computational complexity and hence accelerating the processing time for image segmentation.
  \item With the exception of \cite{SSS}, scale selection methods are limited in their applicability. For example, the majority of the reviewed approaches search for local changes, an approach that might not work with images that consist of many widely homogeneous regions. However, incorporating learning, such as in the approach proposed in this paper and in \cite{SSS} allows to select resolution/scales most suitable for a specific problem at hand (in terms of image modality and the processing task). \parfillskip 0pt plus 0.75\textwidth
  \item The mentioned scale selection approaches propose specific segmentation algorithms that incorporate scale selection. In contrast, the work introduced in this paper is a framework for resolution selection for image segmentation approaches, not tailored toward a specific segmentation method. Therefore, our framework is general and can be used with a wide variety of image segmentation algorithms, such as, level set, graph cuts, region growing, and watershed metods. These segmentation algorithms utilize variable aspects of images other than edges: region homogeneity, textures, colours, and others.
  \item In the proposed approach, the user is given the option of choosing a weighted trade-off between accuracy and speed, a feature that broadens the range of applications for which it can be employed.
\end{itemize}

\section{Proposed Framework} \label{sec:Framework}
In this section we propose a machine-learning framework for automated resolution selection for image segmentation. Determining the best resolution for image segmentation is a difficult task, in part because of the vague definition of the ``best resolution'', which may also differ based on the segmentation method at hand, as a result of the utilization of different aspects of the image (e.g., homogeneity, texture, edges). Hence, we first present a measure for determining the best resolution for an input image when it is segmented with a specific segmentation algorithm.

\begin{figure*}[htb]
\begin{center}
\includegraphics[width=0.7\textwidth]{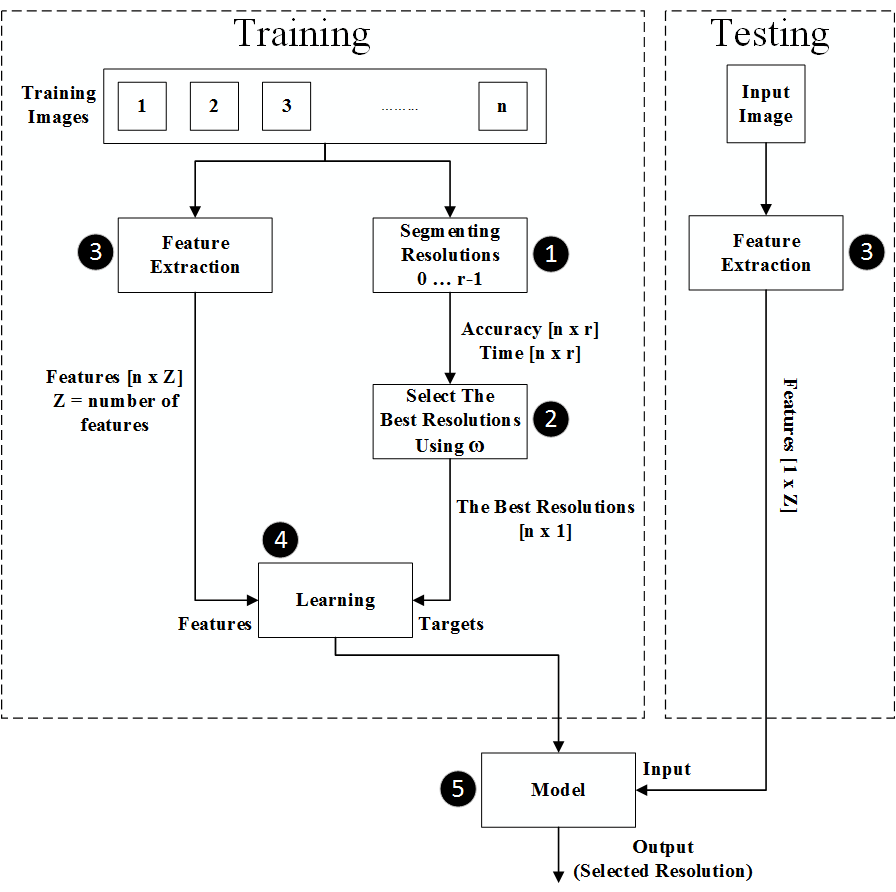}   \\
\caption{Overall framework for the automated selection of the best resolution for image segmentation.}
\label{fig:OverallApproach}
\end{center}
\end{figure*}

The aim of the learning approach, for segmenting an image with a specific algorithm, is to map the features extracted from an input image to the best resolution for image segmentation. Figure \ref{fig:OverallApproach} illustrates the general framework and its two main components: training and testing. In the training phase, the features extracted from training images are assigned to the best resolution for segmenting them. Each training image is segmented at $r$ different resolutions. The accuracy of, and time required for, segmentation at each resolution are recorded, based on which the trade-off measure, $\omega$, is calculated (Eq. \ref{eq:metric}). The best resolution for each image is the one that obtains the maximum value of $\omega$. These best resolutions obtained from the training images are the labels (i.e., classes), which are then used to train the classifier. The inputs are comprised of features extracted from the training images. After training, the classifier is used for estimating the best resolution based on the features extracted from the testing image. The following subsections provide detailed explanations regarding the individual components of the proposed framework.

\subsection{Preliminary Considerations}
For the purpose of this research, a multiresolution representation that includes different subsampling sizes is needed. This requirement excludes the scale-space representation \cite{ScaleSpace}, which has the same spatial resolution for all levels. Although the approximation image of the wavelet representation \cite{Wavelets} can also be used, there would be wavelet information not required for this work (e.g., the high-frequency information defining the horizontal, vertical, and diagonal details images). Therefore, the pyramid representation \cite{Intro-Burt} is chosen for this research. Because of the learning nature of the framework, other multiresolution methods which constructs finite levels of resolutions may also be considered.

The proposed framework is a generic one, in which different feature sets and learning methods can be embedded. Further, the proposed framework can be trained to improve many segmentation algorithms. It cannot be claimed that a set of features or a certain learning technique can always achieve the optimal classification results. The performance of the features varies according to the nature of the  images used, and the learning approach can achieve different results based on several factors, such as data distribution and size. However, the proposed framework may already cover a wide range of image categories and applications, as demonstrated by using diverse sete of test images (see Section \ref{sec:Experiments}).

For this research, many sets of features have been tested, and one of them was selected (Section \ref{sec:featureExtraction}), namely LBP features. Similarly, numerous learning approaches have been investigated, and RAMOBoost (Section \ref{sec:LearningAlg}) achieved the best results.

\subsection{Defining the Best Resolution}  \label{sec:defineOptRes}
The definition of the best resolution for image segmentation can be different from application to application. For example, in fields such as robot navigation, speed is very important, so it could be favoured more than an increase in accuracy. On the other hand, in critical domains, such as medical applications, accuracy is generally much more important since a mistake could be life-threatening. Thus, a method is desired to facilitate decisions about the selection of the best resolution. A measure is thereby proposed that allows the user (or some algorithms) to select a trade-off between speed and accuracy. We use the weighted geometric mean as an aggregation function of several variables \cite{Aggregation1, Aggregation2} and for trade-off measurement \cite{TradeOff}. Given normalized accuracy $A_i$ and time $T_i$ in the range $[0 ~1]$, we define the trade-off measure as
\begin{equation} \label{eq:metric}
\omega_i = A_i^\alpha \times (1-T_i)^{1- \alpha},
\end{equation}
where $i=0,\cdots,r-1$ is the resolution level, and $\alpha \in [0 1]$ can be set by an algorithm or chosen by the user in order to determine the desired trade-off. This trade-off measure is only used during the training phase to define the best (desired) resolutions as labels for the classifier's training, where the segmentation times, $T$, for all resolutions are available. Therefore, $T_i$ can be normalized (i.e., by dividing by the maximum time of all resolutions). Higher values of $\alpha$ will favour accuracy over speed, while lower values would obviously favour speed.

The value of parameter $\alpha$ is set to meet the requirements of the application (to quantify the significance of accuracy and time). Because at the training phase all accuracies and times are available, and due to the limited number of resolutions, the value of $\alpha$ can be quickly found using a simple searching algorithm or by trial-and-error. The selected $\alpha$ is fixed, and the best resolutions are defined accordingly.

The trade-off measure $\omega$ can be defined in several other ways, such as weighted arithmetic mean \cite{Aggregation2} or weighted accuracy over time.

\subsection{Computing the Best Resolutions for Learning} \label{sec:OptResforTrain}
The proposed framework is based on a supervised learning approach for resolution selection for image segmentation. Supervised methods require labeled instances, whereas the labels are the learning targets. In order to learn the best resolutions, they must be provided as labels (i.e., classes) associated with the training dataset (Boxes 1 and 2 in Figure \ref{fig:OverallApproach}). As described in Section \ref{sec:defineOptRes}, the best resolutions can be defined using the trade-off measure, $\omega$. The best resolution is the one that produces the maximum $\omega$. To this end, the image should be segmented at each resolution, and the resulting values for accuracy and time should be recorded.

The pyramid representation entails $r$ resolutions, $0,1,\dots,r-1$, where $0$ denotes the original resolution, and $r-1$ is the coarsest one. The criterion for choosing the lowest level of resolution is the disappearance of image information that is useful for segmentation, which can be found experimentally. In the pyramid representation, each level of resolution is a smoothed and subsampled version of the previous level. After an image is segmented at each resolution and the values for accuracy and time at each resolution are obtained, the trade-ft measure $\omega$ can be calculated for each level of resolution. Considering all $\omega_i$ values, with $i=0,1,\dots, r-1$, the best resolution is the one having the maximum value of $\omega$. This process is described in detail in Algorithm \ref{alg:CalcOptRes}. This method of obtaining the best resolution for image segmentation can be used with a wide range of image segmentation algorithms.

\begin{algorithm}[h!tb]
\caption{Calculating The Best Resolution}
\label{alg:CalcOptRes}
\begin{algorithmic}[1]
\STATE \textbf{Inputs}
\STATE $I$: input image
\STATE $G$: gold standard image (prepared by an expert)
\STATE $r$: number of resolutions
\STATE $\alpha$: trade-off parameter between accuracy and speed for calculating the trade-off measure $\omega$

\STATE $P = pyramid(I,r)$ [pyramid decomposition]
\FOR{i=0,1,\dots,r-1}
\STATE start time calculation
\STATE S = segment($P_i$)
\IF {$i > 0$}
\STATE upS = upSample(S)
\STATE finalS = fineTune(upS)
\ELSE
\STATE finalS = S;
\ENDIF
\STATE $T_i$ = stop time calculation
\STATE $A_i$ = calcAccuracy(finalS, G)
\ENDFOR

\FOR{i=0,1,\dots,r-1}
\STATE $\omega_i = A_i^\alpha \times \left( 1 - \frac{T_i}{\underset{k}{\text{max}}~ T_k}\right)^{1-\alpha}$
\ENDFOR

\STATE BestRes = $i \, \mid \,\omega_i = \underset{j}{\text{max}}\, \omega_j$, $j=0,1,\cdots, r-1$
\end{algorithmic}
\end{algorithm}

Global resolution selection will not always results in best accuracies for all regions in the image. Therefore, after segmentation at lower resolutions and the upsampling of the results to the original resolution, fast fine-tuning is needed in order to compensate the loss in quality (Algorithm \ref{alg:CalcOptRes}, line 12). The literature reports numerous fine-tuning methods, as discussed in Section \ref{sec:MultiResImgSeg}, but the majority are slow, and each method is designed for a particular segmentation algorithm. This work applies the region growing approach for fine-tuning, which is appropriate for many segmentation algorithm. it starts from the border of the results to perform the fine-tuning. Region growing was selected for the following reasons:
\begin{itemize}
  \item It can be used as a boundary-enhancing method following many other segmentation techniques.
  \item It is simple and very fast, an attribute that translates into only minimal increase in computation time.
  \item It is consistent and predictable in its time requirement, which enables the learning method to learn the best resolution. In contrast, some methods result in an unpredictable increase in time, which makes the learning impossible, as illustrated in the forthcoming example.
\end{itemize}

\begin{figure}[htb]
\begin{center}
\begin{tabular}{ccc}
\includegraphics[height=4.3cm]{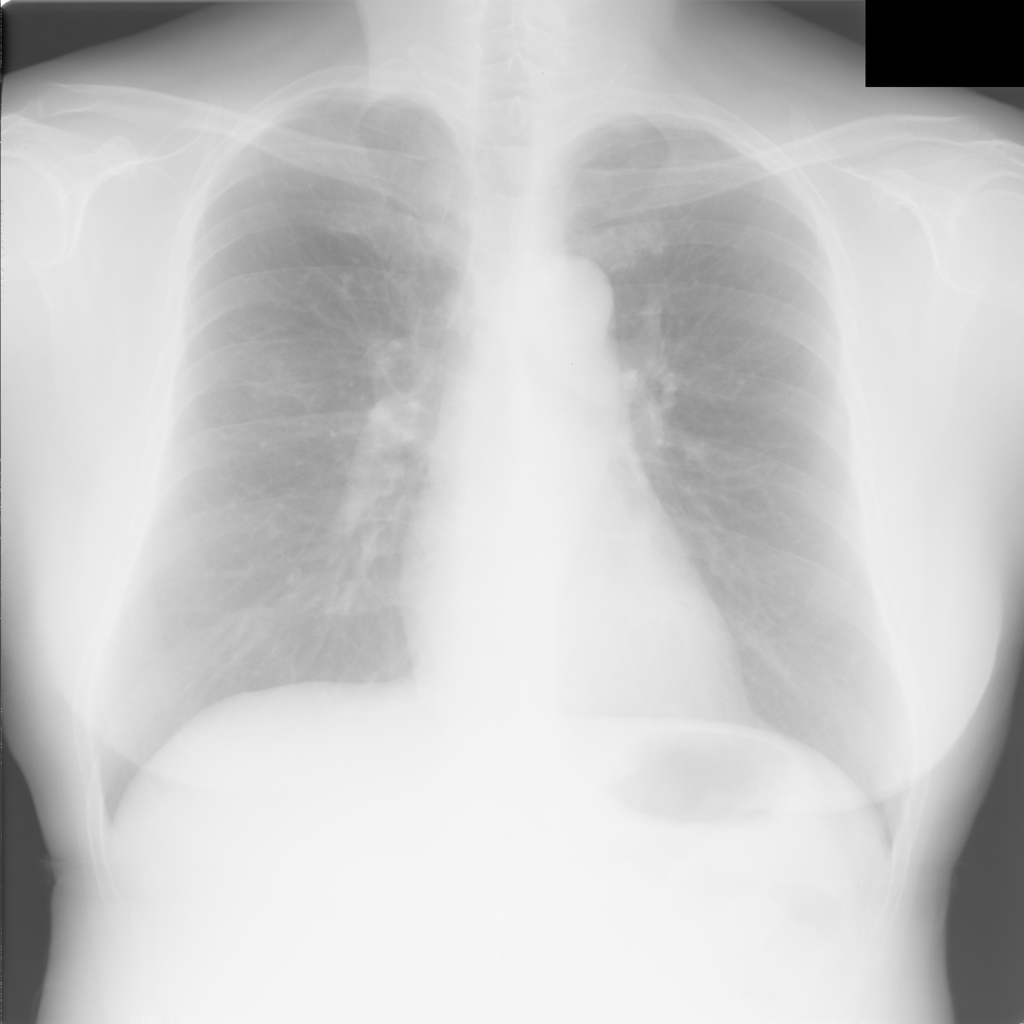} &
\includegraphics[height=4.3cm]{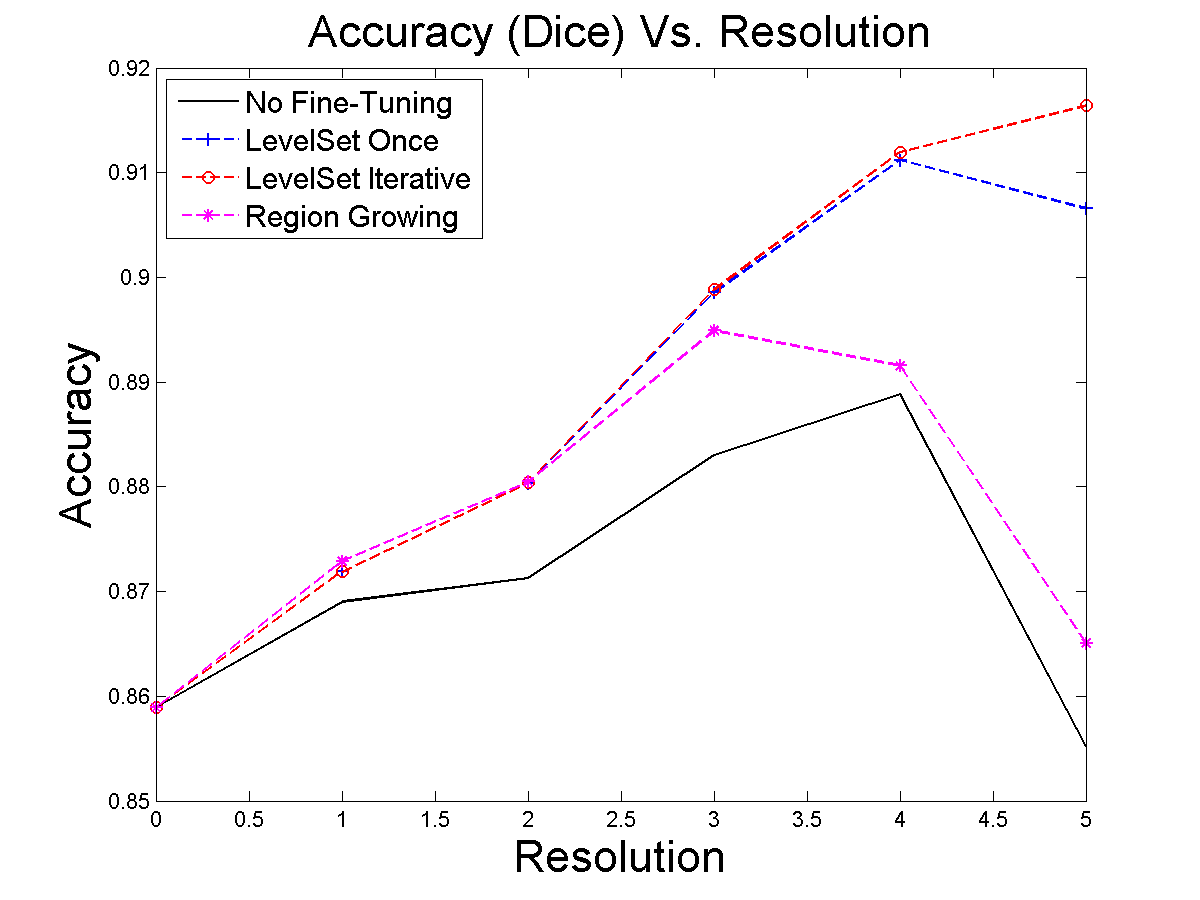} &
\includegraphics[height=4.3cm]{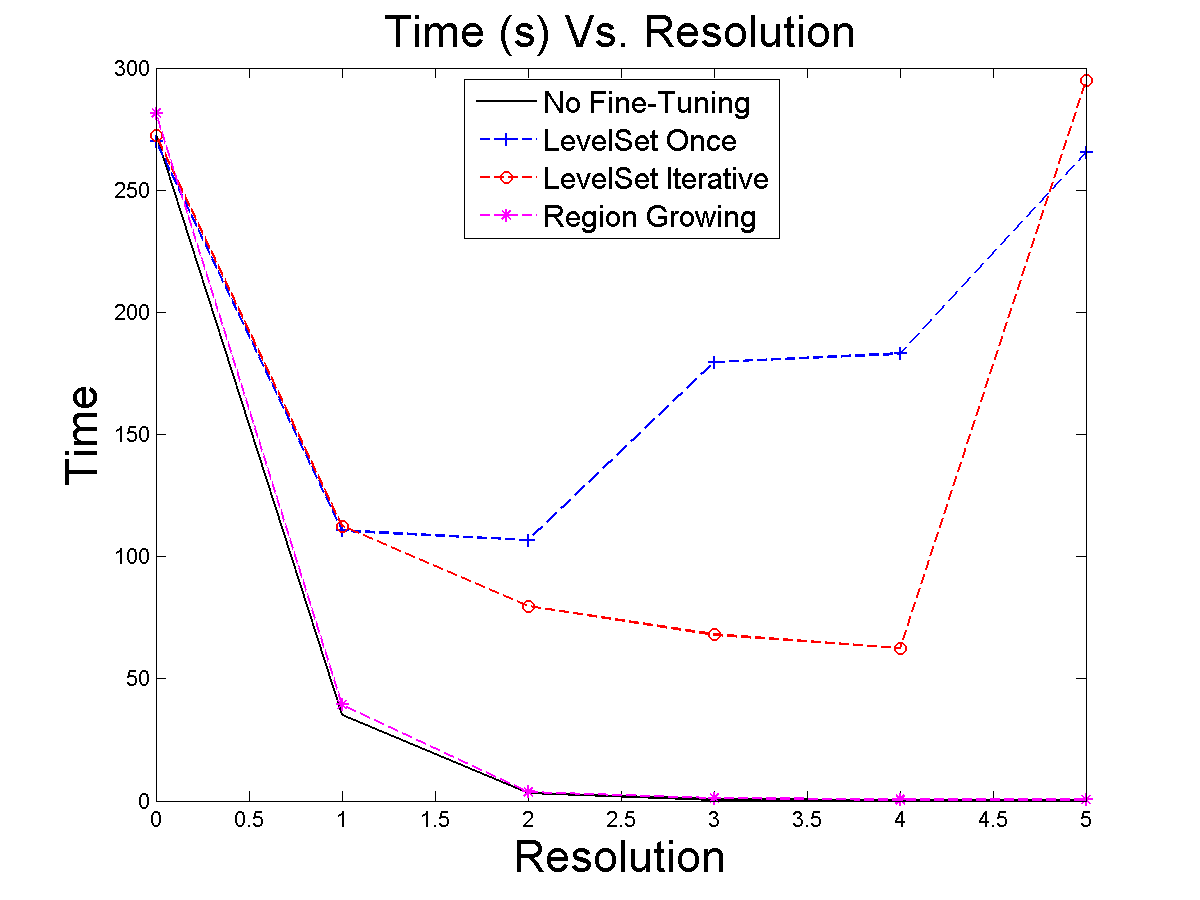}
\end{tabular}
\caption{An image segmented using ChanVese level set and fine-tuned with level set once at the original resolution, iteratively at each resolution, region growing at the original resolution, and without fine-tuning. From top to bottom: input image, accuracies, and times.}
\label{fig:RegionGrowingJustify}
\end{center}
\end{figure}

Figure \ref{fig:RegionGrowingJustify} shows an image segmented using the ChanVese level set \cite{regionLevelSets-ChanVese}. The accuracy (dice coefficient, Eq. \ref{eq:dice}) and time in seconds are presented for the following four cases: no fine-tuning, fine-tuning with ChanVese only at the original resolution, fine-tuning with ChanVese iteratively in each resolution, and fine-tuning with region growing at the original resolution. Although fine-tuning with ChanVese results in a better accuracy than region growing, it results in a completely unpredictable increase in time. On the other hand, region growing results in a consistent and slight increase in time.

\subsection{Features Extraction} \label{sec:featureExtraction}
The extraction of relevant features is very important for achieving reasonable classification accuracy (Box 3 in Figure \ref{fig:OverallApproach}). Two criteria were set for the choice of feature extraction methods in this research: good correlation with the classes (resolutions) and  the time constraint. The performance of the features in terms of classification accuracy was assessed with the $F_1$\emph{-measure} described in Section \ref{sec:PerfMeasures}. The speed was measured in seconds. Relevant features must be extracted extremely fast so as not to delay the decision about the best resolution. From a set of good features, several of them have therefore been omitted because they would slow down the process. Examples of such features are Gabor \cite{GaborFeatures} and Granulometric features \cite{GranulometricFeatures}. We have used Local Binary Patterns (LBP) in this research because they can be  extracted fast and obtain good accuracy.

\subsubsection{Local Binary Patterns (LBP)}
LBP is a method for extracting texture features proposed by Ojala et al. \cite{LBP1}. It was originally a 3$\times$3 operator, where the intensities inside each block are thresholded by the value of the central pixel. The value of the central pixel is then obtained from the summation of the thresholded values multiplied by powers of two. In this way, $2^8=256$ different values can be achieved. Recently, Radon barcodes have been introduced that could be used as an alternative to LBP \cite{tizhoosh2015,tizhoosh2016}.

Constructing features from the histogram of an LBP-labeled image has been proven to be efficient \cite{LBP-Face1,LBP-Face2}. In order to retain spatial information, the image is divided into $m$ regions $R_0, R_1, \cdots, R_{m-1}$, in which the histogram features are calculated as follows \cite{LBP-Face2}:

\begin{equation}
\begin{split}
H_{i,j} = \sum_{x,y} \Gamma\{I_{LBP}(x,y)=i\} \Gamma\{(x,y) \in R_j\}, \\ i=0,\cdots,n-1, j=0,\cdots,m-1,
\end{split}
\end{equation}
where $I_{LBP}$ is an LBP labeled image, $n$ is the number of labels produced by the LBP operator, and $\Gamma\{A\}$ is defined as
\begin{equation}
\Gamma\{A\} = \left\lbrace
\begin{array}{l l}
1 & \quad \mbox{if $A$ is True},\\
0 & \quad \mbox{if $A$ is False}.\\
\end{array} \right.
\end{equation}

LBP is simple and very fast, making it appropriate for the purposes of this research. LBP is also robust to monotonic grey-scale changes. Several LBP variants have been proposed, such as larger operator size \cite{LBP2}, rotation invariance \cite{LBP2,LBP3}, and multiscale LBP \cite{MultiScaleLBP}.

\subsection{Learning the Best Resolution} \label{sec:learnOptRes}
Objects with different characteristics (e.g. shape, size, texture) may be segmented more accurately at specific resolutions, as can be observed from the experiments described in Section \ref{sec:propMeasurePerf} and as well as demonstrated in \cite{SegDiffRes1, SegDiffRes2}. Coarse resolutions are generally sufficient for segmenting large objects with clear boundaries. This is not the case with small objects, fine details, and sharp corners. Other factors may affect the choice of the best resolution, such as noise, which can be suppressed as the resolution decreases, allowing more accurate and faster segmentation. Unfortunately, natural images (non-synthetic) are rather complex. Most of the time, the same image might have a variety of noise levels and contain objects with different characteristics. Resolution selection for image segmentation is thus difficult. In this work, a machine-learning approach is proposed (Boxes 4 and 5 in Figure \ref{fig:OverallApproach}), so that best resolutions can be learned based on features extracted from the training images.

When the features from the images are already available (Section \ref{sec:featureExtraction}), as well as the training instance labels (Section \ref{sec:OptResforTrain}), a supervised learning approach can be used to learn the best resolution. Learning the best resolutions (offline tarinign) is explained in Algorithm \ref{alg:OptimalResLearning}, and estimating the best resolution for an input image (online autmation) is presented in Algorithm \ref{alg:OptimalResEstimation}.

\begin{algorithm}[htb]
\caption{Offline training to learn the best resolutions}
\label{alg:OptimalResLearning}
\begin{algorithmic}[1]
\STATE Input: Training images $I_j$, $j=1,2,\cdots,n$
\STATE Input: Corresponding gold standard images $G_j$
\STATE Set $r$ (number of resolutions)
\STATE Set $\alpha$ (the parameter for the trade-off measure $\omega(\alpha)$)

\STATE \textbf{Initialize:} inputs = [ ], targets = [ ]
\FOR{i=1,2,\dots,n}
\STATE BestRes = Algorithm\ref{alg:CalcOptRes}($I_i, G_i, r, \alpha$)
\STATE $F$ = featuresExtraction($I_i$)
\STATE inputs = append(inputs, $F$)
\STATE targets = append(targets, BestRes)
\ENDFOR

\STATE model = trainClassifier(inputs, targets)
\STATE Save $LM$ = model
\end{algorithmic}
\end{algorithm}

\begin{algorithm}
\caption{Online automation}
\label{alg:OptimalResEstimation}
\begin{algorithmic}
\STATE Input: $I$: new (unseen) image
\STATE Input $LM$: read the learned model

\STATE Extract features: $F$ = featuresExtraction($I$)
\STATE Estimate the best resolution: $\hat{r}$ = estimate($LM$, $F$)
\STATE Perform multiresolution segmentation with $\hat{r}$
\end{algorithmic}
\end{algorithm}

The learning objective here is to estimate the best resolution for segmenting an input image using the extracted features. As shown in Section \ref{sec:propMeasurePerf}, the learning data in this research have an imbalanced class distribution. This problem is well recognized in the machine learning community, and it can seriously affect learning performance \cite{Imbalanced_review, Imbalanced_He}. Therefore, a learning method specifically designed to learn from imbalanced data was used.

\subsubsection{The Learning Algorithm} \label{sec:LearningAlg}
We used RAMOBoost \cite{RAMOBoost} for learning and estimating the best resolution. RAMOBoost is a variant of AdaBoost that is specifically designed for learning from imbalanced data through adaptive generation of synthetic samples of minority class examples in each iteration of the AdaBoost method. RAMOBoost has been chosen because of its good performance, which stems from its two combined components boosting and sampling:

\begin{itemize}
  \item Boosting -- trains several base classifiers consecutively while adaptively adjusting the weights of the training instances. In this manner, the decision boundary is adaptively shifted during each boosting iteration, so that the focus is on instances that are difficult to learn. This property provides ability to deal with outliers \cite{AdaBoost_Outliers}. The generalization abilities of boosting methods have been proven by Schapire and Freund \cite{BoostingBook}.
  \item Sampling -- in RAMOBoost, the sampling is based on adaptive adjustment to the sampling weights of minority class samples according to their distribution. Greater emphasis is thus placed on rare examples that are inherently difficult to learn.
\end{itemize}

Decision trees are selected as the base classifier for RAMOBoost in this research. It is the most popular choice as the base classifier for AdaBoost \cite{AdaBoost_DT1}. An experimental comparison of the performance of decision trees against Naive Bayes and Bayes Net as base classifiers is presented in \cite{AdaBoost_DT_Comp}. The authors reported that AdaBoost with decision trees as base classifier achieved the highest classification rate with lowest computational time.

The model resulting from training with RAMOBoost consists of $n_t$ decision trees classifiers, where $n_t$ is the number of iterations. Classification based on this model is performed by finding the class that maximizes a weighted average of the outputs of all base classifiers.

\subsection{Performance Measures} \label{sec:PerfMeasures}
Measuring classifiers' performance is an important issue in machine learning \cite{EvalBook}. A classifier is typically evaluated using a confusion matrix 
Elements of the confusion matrix count the number of estimated classes with respect to actual classes. The diagonal elements represent correctly classified instances. The most widely used measure is accuracy, which is calculated from the confusion matrix as follows:
\begin{equation} \label{eq:accuracy}
\textrm{Accuracy} = \frac{\sum_{i=1}^n a_{ii}}{\sum_{i=1}^n \sum_{j=1}^n a_{ij}}.
\end{equation}


\noindent However, the level of accuracy can be misleading when used for imbalanced class problems \cite{Imbalanced_He}. For example, given a data with 2\% of minority class instances and 98\% of majority class instances, an accuracy of 98\% can be achieved by blindly classifying all the data instances as the majority class. The literature includes descriptions of several measures for evaluating the performance of the classification of imbalanced class problems \cite{Imbalanced_He,Imbalanced_review,Gmean}. For class $C_i$, two measures, precision ($P_i$) and recall ($R_i$), can be calculated from the confusion matrix as follows:

\begin{equation} \label{eq:precision}
P_i = \frac{a_{ii}}{\sum_{j=1}^n a_{ji}},
\end{equation}

and

\begin{equation} \label{eq:recall}
R_i = \frac{a_{ii}}{\sum_{i=1}^n a_{ij}}.
\end{equation} \label{eq:FMeasure}

$F_1$\emph{-measure} is the harmonic mean of precision and recall, which is calculated by

\begin{equation}
F_1\textrm{-measure} = \frac{2 P_i R_i}{P_i + R_i}.
\end{equation}

$G$\emph{-mean} is the geometric mean of precision and recall. It has been extended for multiclass problem evaluation by Sun et al. \cite{Gmean} and has been defined as

\begin{equation} \label{eq:Gmean}
G\textrm{-mean} = \left(\prod_{i=1}^n R_i\right)^{1/n},
\end{equation}

where $n$ is the number of classes. However, if any class has 0 recall, the $G$\emph{-mean} will also be 0, which would be misleading as an overall performance indication.

The Area Under Curve (AUC), using the Receiver Operating Characteristic (ROC) curve, is a well-known method for evaluating a classifier's performance as well. It is useful for evaluating imbalanced class problems \cite{Imbalanced_He,Imbalanced_review}. AUC measures the overall performance of a classifier. However, it cannot represent the performance of different parts of the ROC curve \cite{ROCcurve}. Curves in ROC space might intersect with each other; therefore, classifiers with high AUC value may have worse performance at some regions of ROC space than a classifier with lower value of AUC.

In this research, the $F_1$\emph{-measure} was used for evaluating the performance of the estimation of the best resolutions for image segmentation. Unlike accuracy (Eq. \ref{eq:accuracy}), $F_1$\emph{-measure}, which is a weighted harmonic mean of precision (Eq. \ref{eq:precision}) and recall (Eq. \ref{eq:recall}), is more useful for problems with imbalanced class distribution. By using $F_1$\emph{-measure}, the performance of classification per class (i.e., resolution) can be assessed. This enables the investigation of the performance of classification of individual classes, in contrast to $G$\emph{-mean} (Eq. \ref{eq:Gmean}) and AUC, which only measure the overall performance.

\section{Experimental Results} \label{sec:Experiments}
In this section, the results of experiments conducted to verify the performance of the trade-off measure and resolution selection will be reported. First, settings of the experiments, including datasets and parameters, are explained. Then, the performance of the trade-off measure is discussed. After that, RAMOBoost resolution classification is evaluated. Last subsection discusses the impact of misclassification of resolution on both accuracy and speed.

\subsection{Experimental Setup}
\subsubsection{Datasets}
\begin{figure*}[htb]
\begin{center}
\begin{tabular}{cccc}
\includegraphics[height=4.6cm]{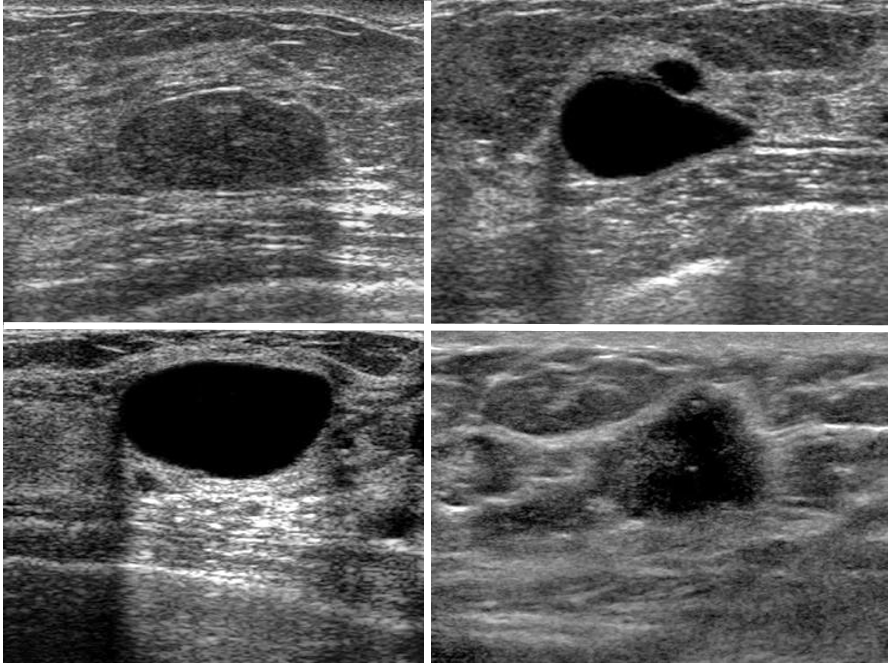} & 
\includegraphics[height=4.6cm]{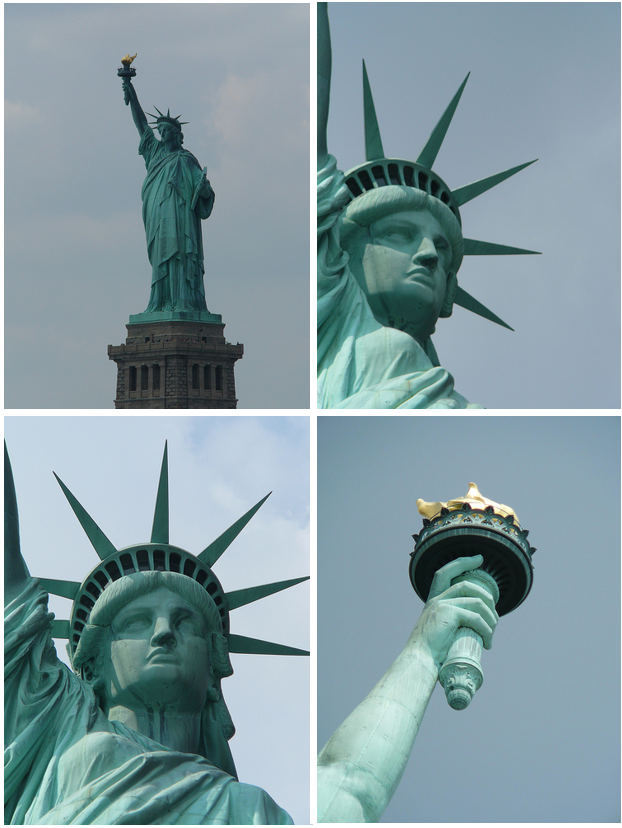} &
\includegraphics[height=4.6cm]{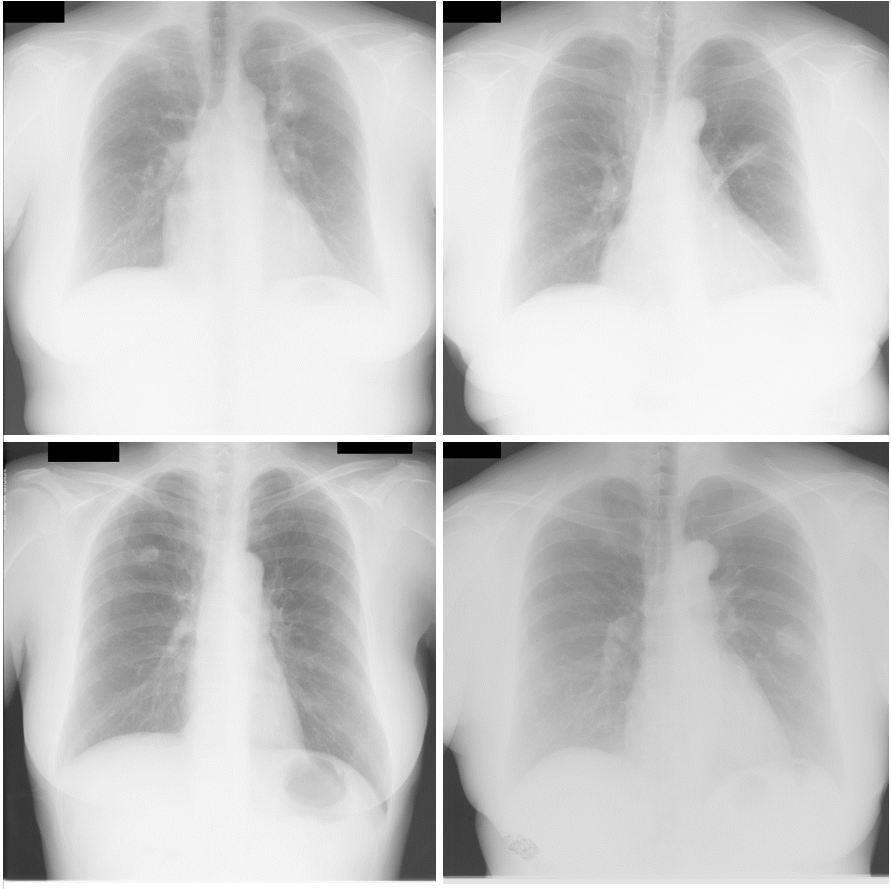}  &   \\
\end{tabular}
\caption{Sample images (left to right): breast ultrasound, Liberty from Cornell iCoseg, and lung X-ray.}
\label{fig:DataSetSamples}
\end{center}
\end{figure*}

Three different image datasets were used for the experimental verification of performance of the proposed resolution selection framework:
\begin{itemize}
\item Breast ultrasound dataset consists of 52 breast ultrasound images whose sizes range from 230$\times$390 pixels to 580$\times$760 pixels. The level of shape variability in the images is high. The presence of speckle noise and low local contrast make these images difficult to segment. Segmentation algorithms for this kind of images usually require specially designed preprocessing and/or post-processing filters. Because the objective of this research was not to design an image segmentation method for a particular dataset, a semi-automated approach was applied, whereby the user selects one of the several output segments by clicking on it. This has been automated by taking the centroid of the object in the gold standard image simulating the user's click.
\item Liberty statue (Liberty) from CMU-Cornell iCoseg image segmentation dataset \cite{iCosegDataset} consists of 41 images with a high degree of shape/scale variability. Some images contain the whole statue, while others contain only part of it. The size of the images is 375$\times$500 pixels.
\item Lung X-ray dataset \cite{LungXRayDataset} consists of 98 lung X-ray images that are of 1024$\times$1024 pixels in size. Compared to the images in the previous two datasets, these images exhibit a high degree of similarity. The contrast between the objects (lungs) and the background is low.
\end{itemize}
Samples from these datasets are presented in Figure \ref{fig:DataSetSamples}.

\subsubsection{Parameter Settings}
Obtaining the best resolutions as labels for training requires the images to be segmented and evaluated at all possible resolutions. For the datasets used, it was found that after the sixth level of resolution, no meaningful information remained in the image. The pyramid representation was thus comprised of six resolutions (i.e., $r$ = 6), where 0 is the original image resolution, and 5 is the coarsest one. The pyramid was implemented as described in \cite{Intro-Burt}. A separable 5$\times$5 filter, $w$, was employed, defined by

\begin{equation}
w = [\frac{1}{4}-\frac{a}{2} , \frac{1}{4} , a , \frac{1}{4} , \frac{1}{4}-\frac{a}{2}].
\end{equation}

Parameter $a$ was selected as 0.4, so that the filter would be close to a Gaussian shape \cite{Intro-Burt}.

The purpose of this research was to select the best resolution for the segmentation of an input image, and not segmentation itself, including optimal parameter selection, preprocessing, or post-processing. The parameters selected for the employed image segmentation algorithms might therefore not be the optimal choice for the respective datasets. However, some images, such as breast ultrasound and lung X-ray images, are difficult to segment without preprocessing or post-processing. For the breast ultrasound dataset, a semi-automated segmentation approach was used as described in the previous subsection. Because of the low contrast in the lung X-ray dataset images, their contrast was enhanced through contrast-limited adaptive histogram equalization \cite{preProc-Contr}.

In order to investigate the resolution selection framework with different image segmentation algorithms, two well-known methods were selected. They were chosen to have different concepts behind them. Parametric Kernel Graph Cuts (PKGraphCuts) \cite{ParKernelGraphCuts} is based on graph theory, and ChanVese level set \cite{regionLevelSets-ChanVese} is an active contour method. PKGraphCuts, implemented by Ben Ayed, was utilized in this research (code available at \cite{PKGraphCutsCode}). The regularization weight parameter $\alpha$ was set to 0.1. The initialization of the ChanVese level set was performed as multiple circles, which have been shown to be effective \cite{regionLevelSets-ChanVese}. The iterations stop if a change less than $\eta$ (= 5) occurs for five consecutive iterations, or if the number of iterations exceeds a preset threshold value (here 1000).

Images of each dataset were segmented using different segmentation algorithms. The choice of segmentation algorithm for each dataset was performed empirically based on random samples of images from each dataset:
\begin{itemize}
  \item Breast ultrasound dataset: PKGraphCuts
  \item Liberty dataset: PKGraphCuts
  \item X-ray Lung dataset: ChanVese level set
\end{itemize}

The accuracy of the segmentation is measured using dice coefficient defined as follows \cite{ER-Dice}:

\begin{equation} \label{eq:dice}
\textrm{Dice coefficient} = \frac{2\vert I_n \cap I_G \vert}{\vert I_n \vert + \vert I_G \vert},
\end{equation}

\noindent where $I_n$ is the nth segmented image, $I_G$ is its gold standard image, and $\vert \cdot \vert$ indicates the set cardinality.

In the experiments conducted for this research, the performance achieved with RAMOBoost was compared with that obtained with AdaBoost, Support Vector Machines, and SVM with re-sampled training data using adaptive synthetic sampling (ADASYN) \cite{ADASYN} (SVM-ADASYN). LibSVM \cite{libSVM} (code available at \cite{LibSVMCode}) implementation of SVM was used, with a radial basis function for the kernel. The penalty parameter $C$ and the parameter $\gamma$ for radial basis function were selected through 5-fold cross-validation. Joint Mutual Information (JMI) feature selection \cite{YangMoody} was used for the selection of the 10 most representative features for training the SVM. For ADASYN, the number of nearest neighbours was 5, and the balance level parameter $\beta$ was chosen as 0.7. For both AdaBoost and RAMOBoost, the decision tree classifier was used as the base classifier. Both boosting algorithms were run for 10 iterations. LBP features were extracted using 10-bins histogram for each region. For all methods, the training and testing sets were split by 10-fold cross-validation. Each experiment was run 10 times, and the average was taken.

\subsubsection{Implementation Environment}
The experiments were conducted using a PC with a CPU speed of 2.20 GHz and 8 GB of RAM. The operating system was Windows 7 (64-bit version). The program was written and run with 64-bit version of Matlab$\texttrademark$.

\subsection{Trade-off Measure Performance} \label{sec:propMeasurePerf}
\begin{figure*}[htb]
\begin{center}
\begin{tabular}{ccccc}
\includegraphics[width=2.8cm]{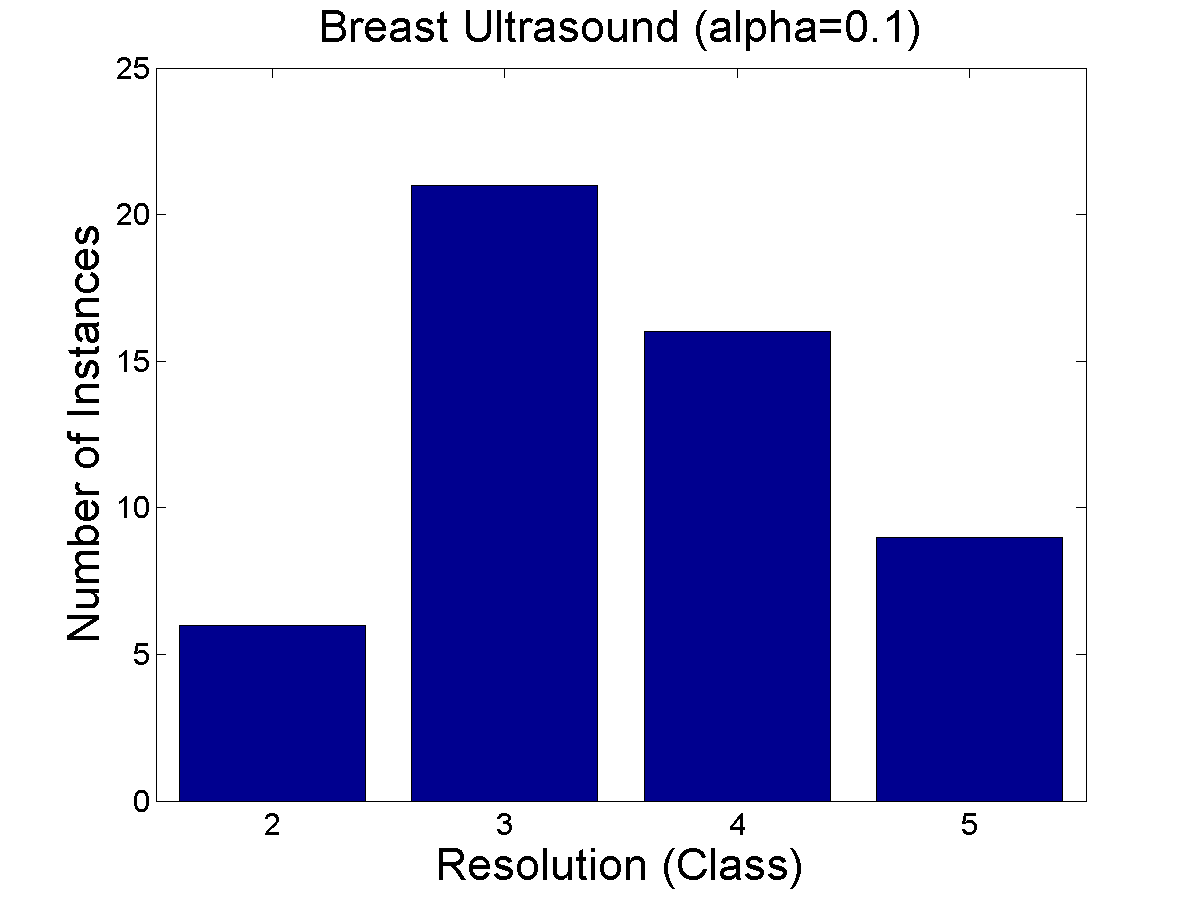} & \includegraphics[width=2.8cm]{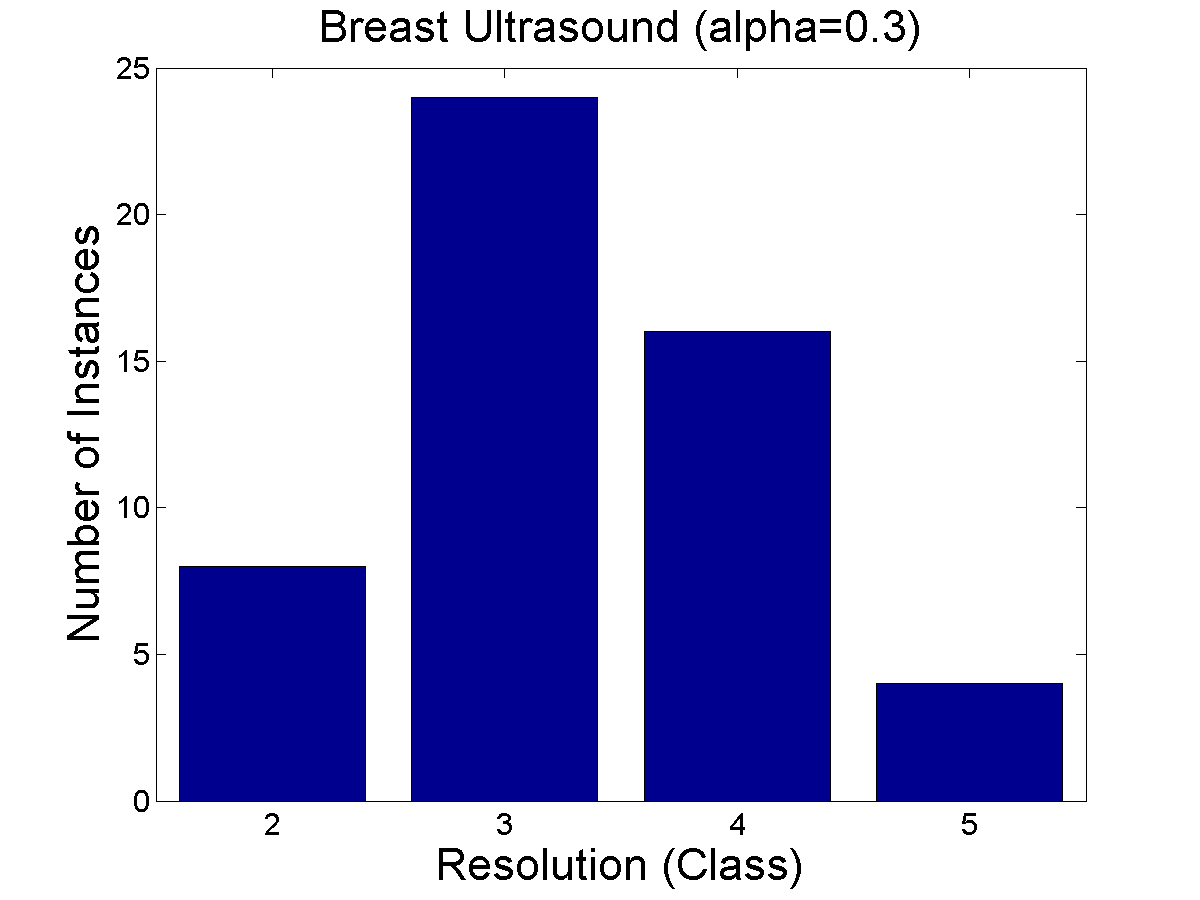} & \includegraphics[width=2.8cm]{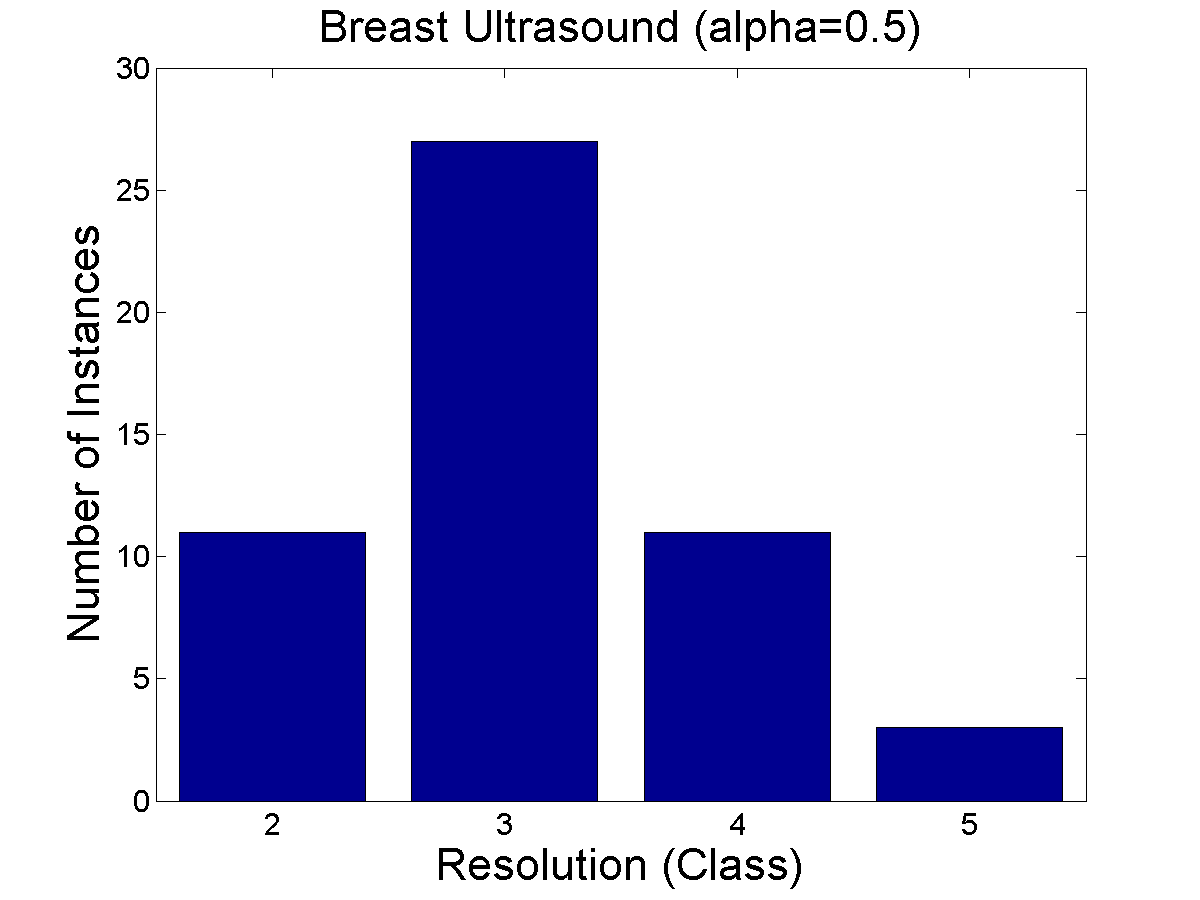} & \includegraphics[width=2.8cm]{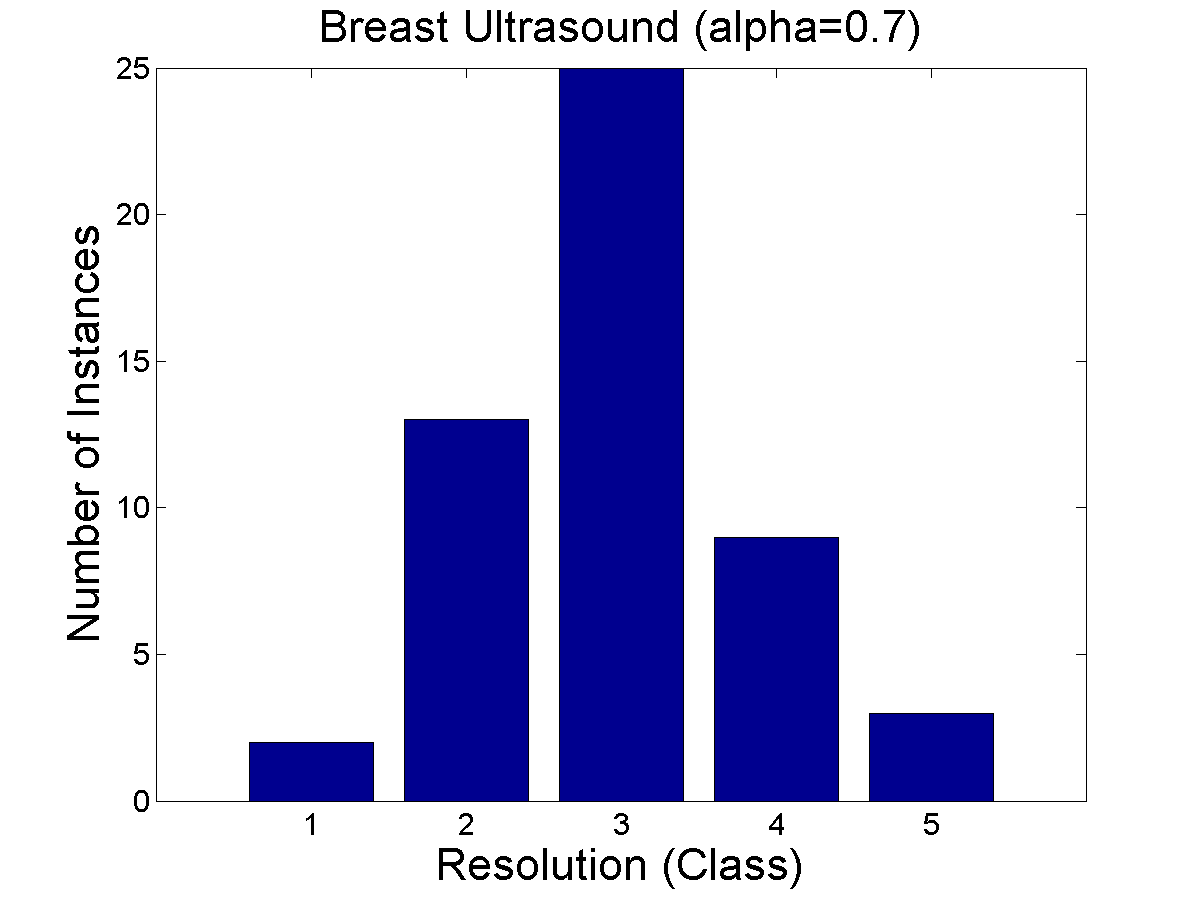} & \includegraphics[width=2.8cm]{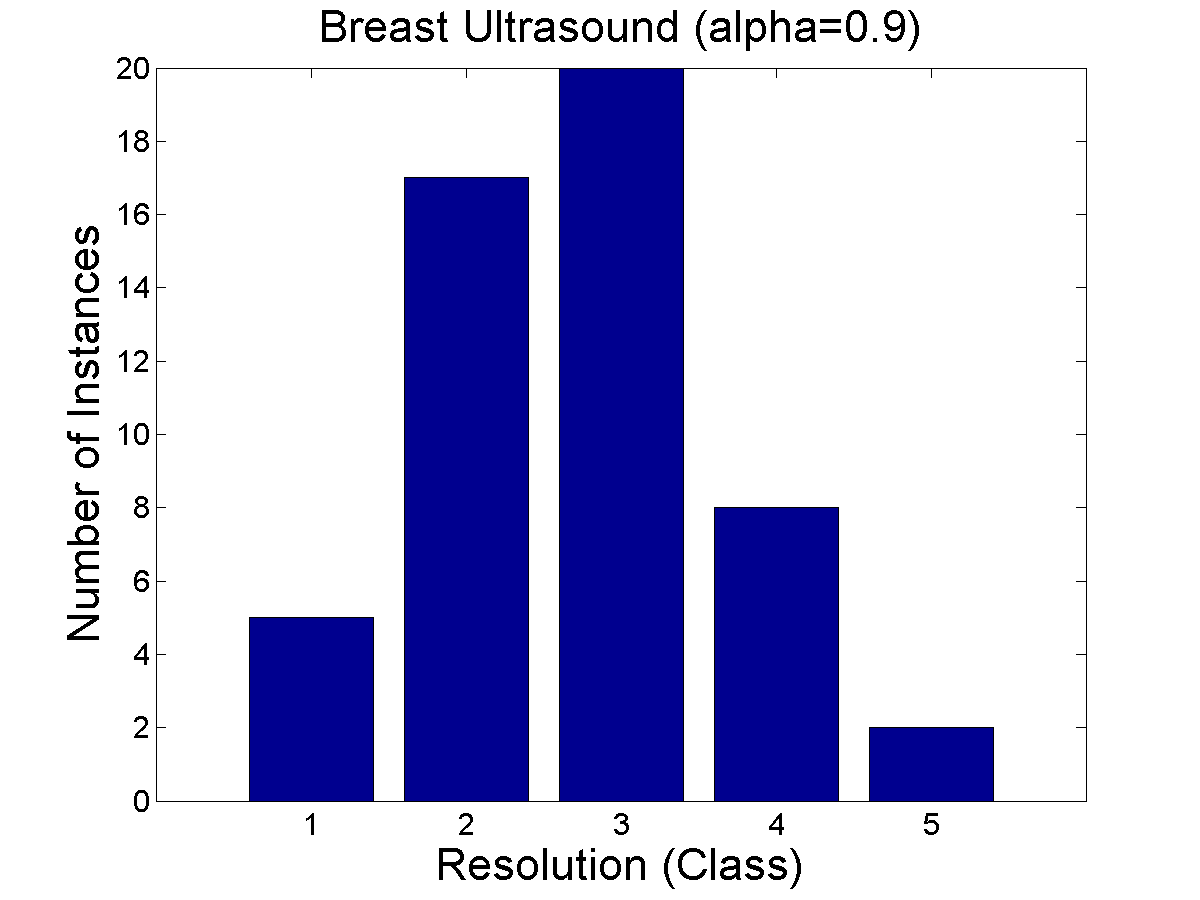}\\

\includegraphics[width=2.9cm]{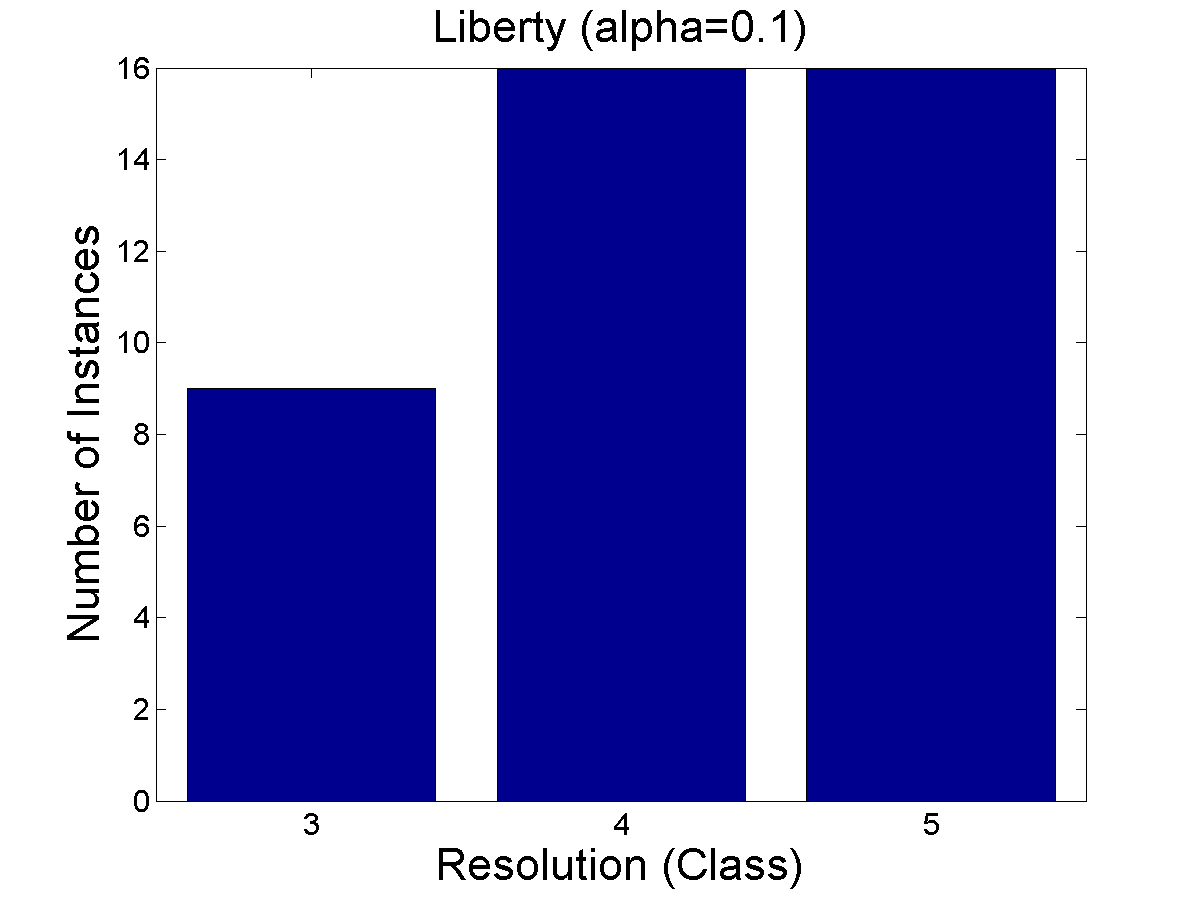} & \includegraphics[width=2.9cm]{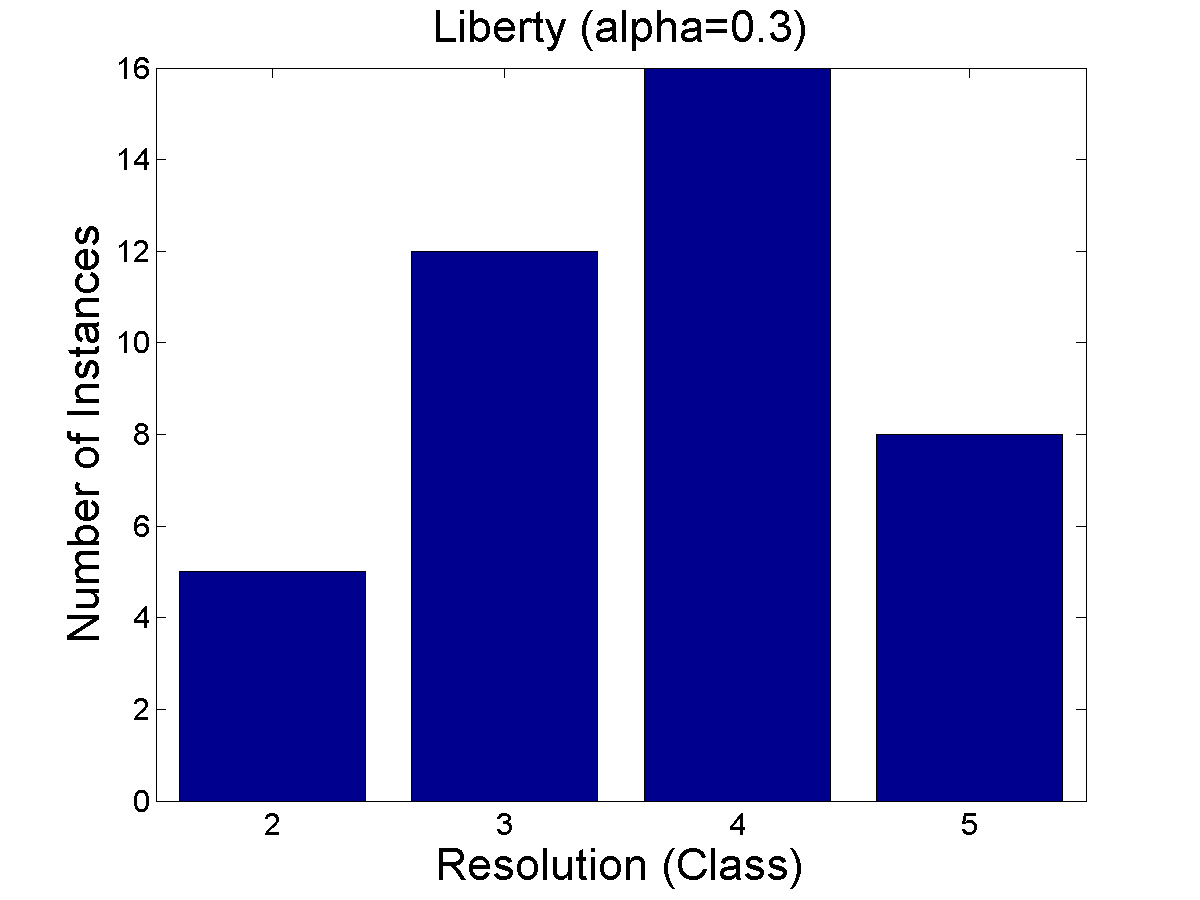} & \includegraphics[width=2.9cm]{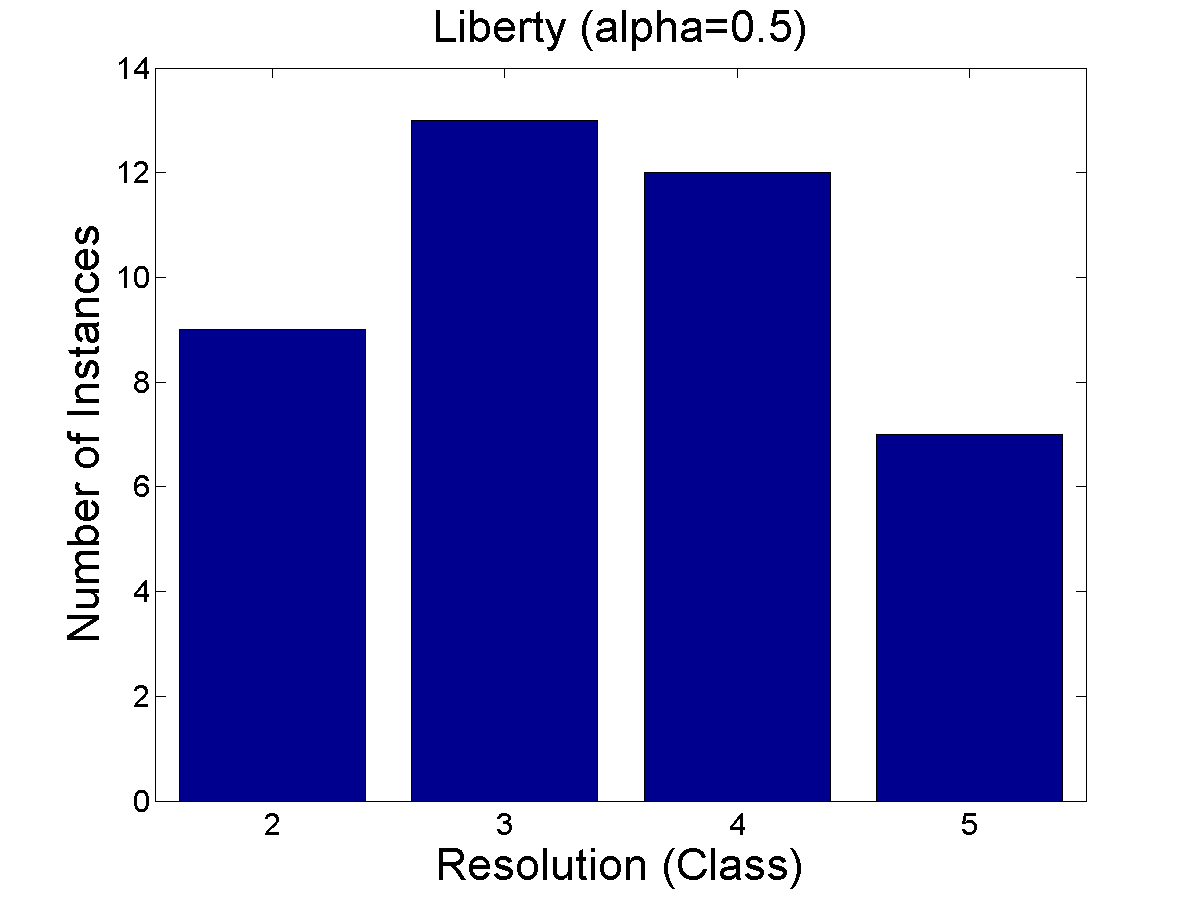} & \includegraphics[width=2.9cm]{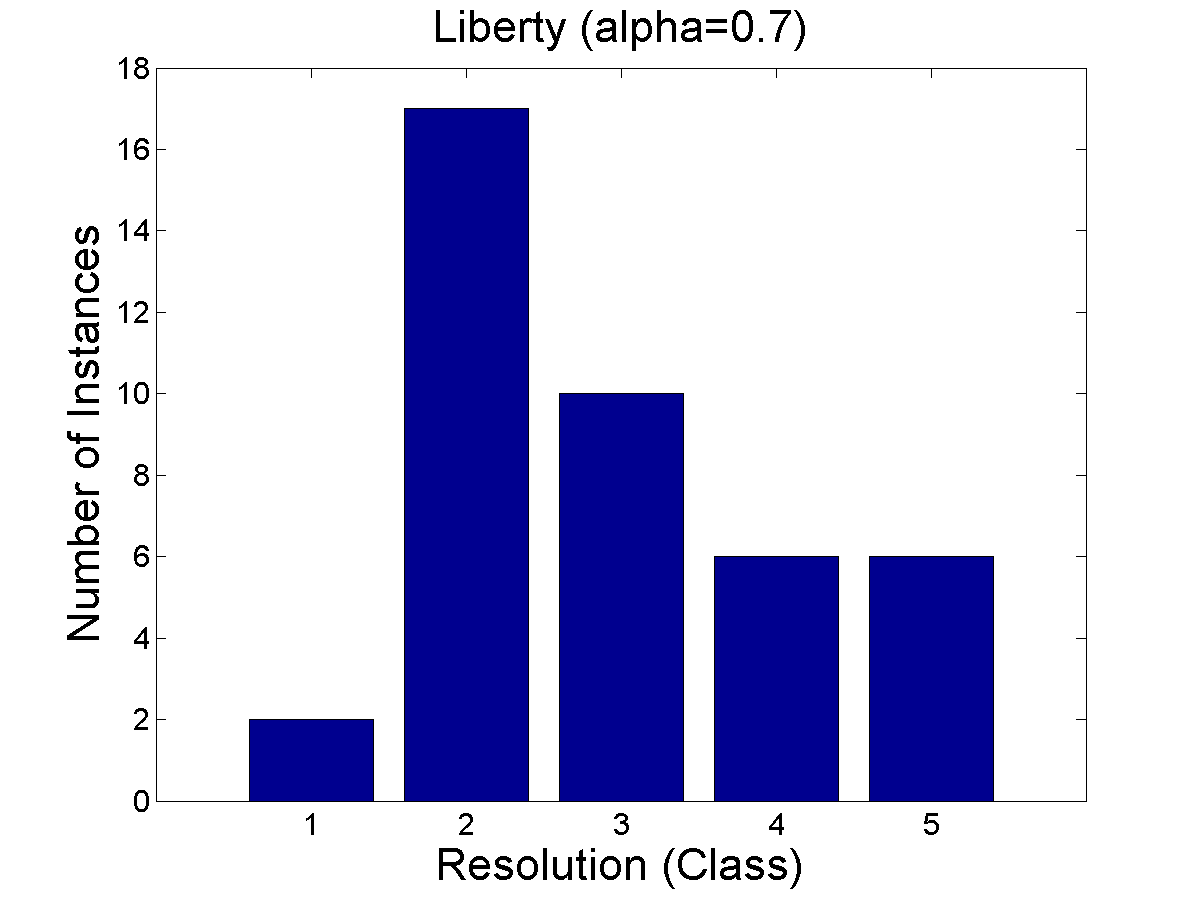} & \includegraphics[width=2.9cm]{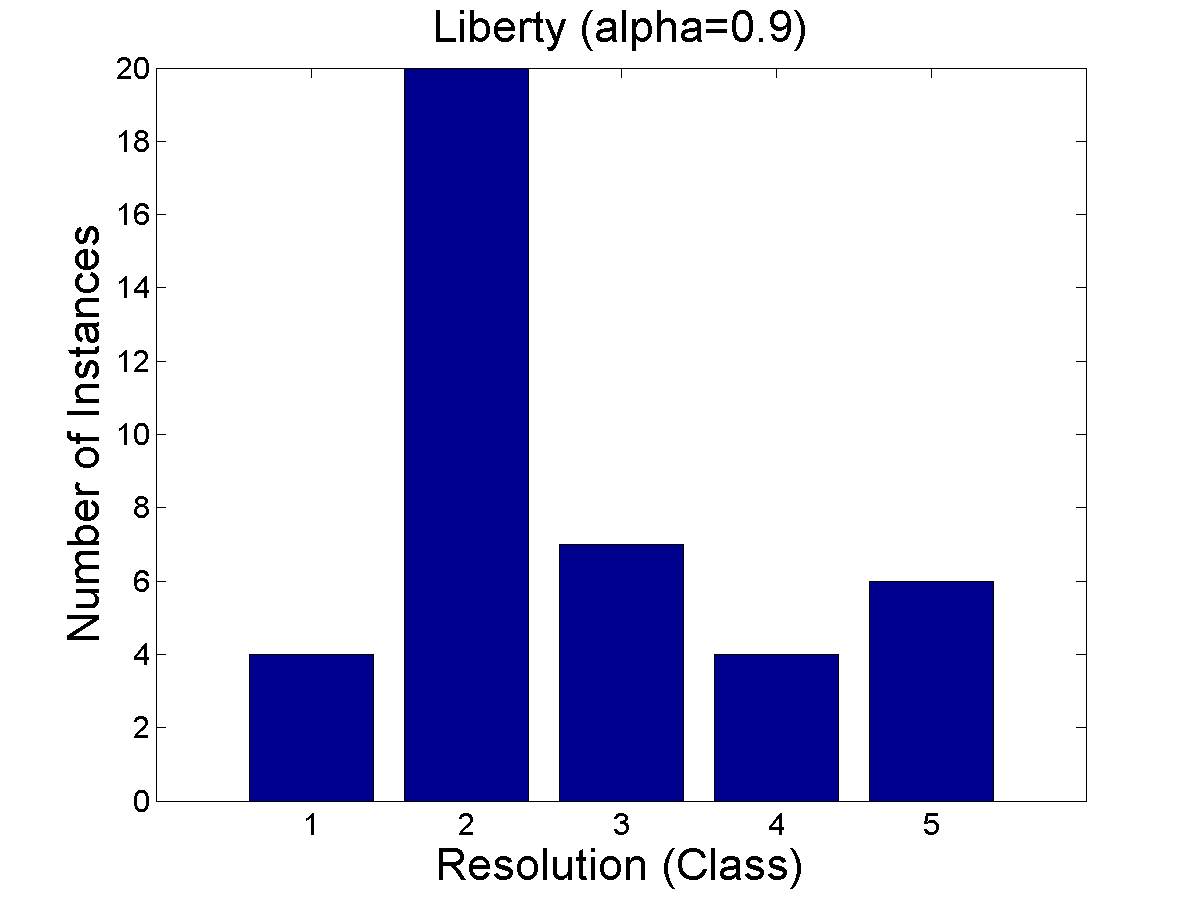}\\

\includegraphics[width=2.9cm]{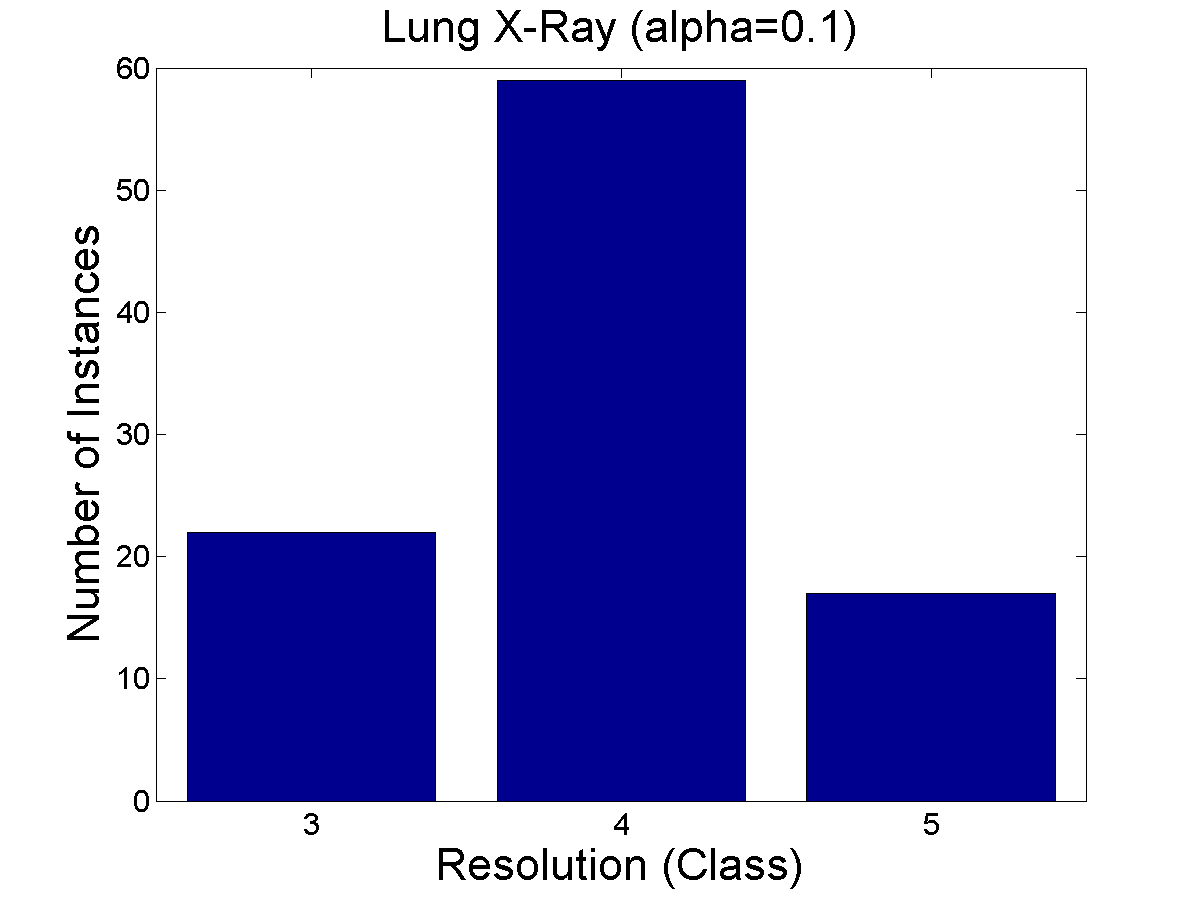} & 
\includegraphics[width=2.9cm]{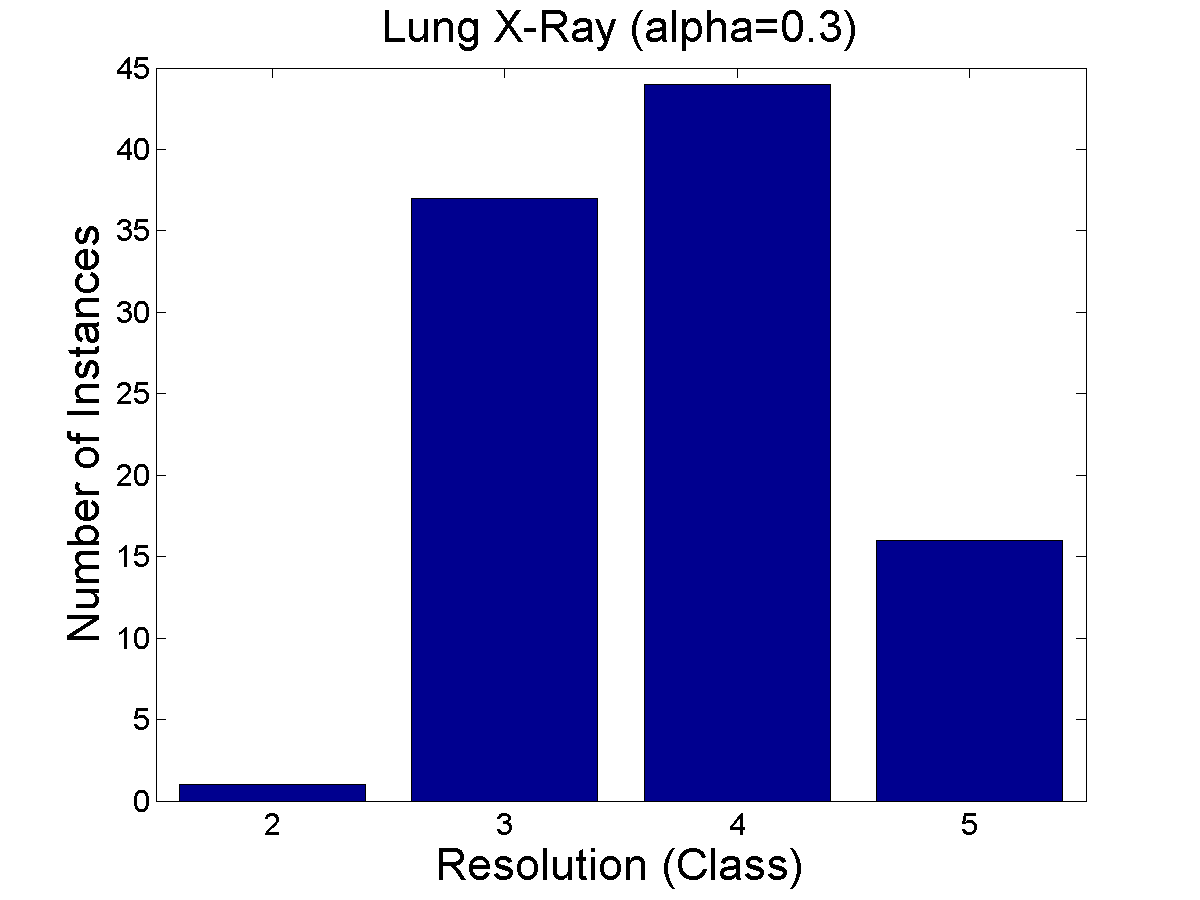} & 
\includegraphics[width=2.9cm]{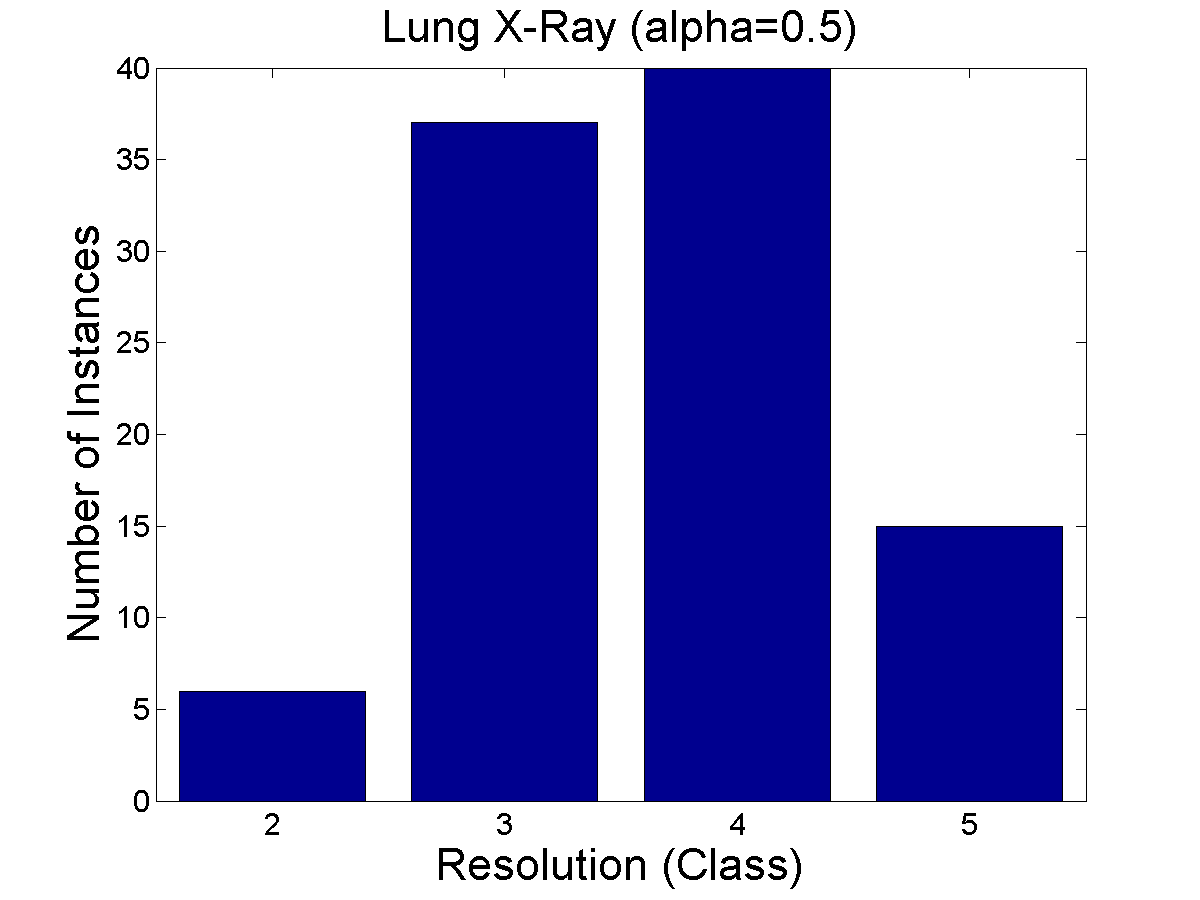} & 
\includegraphics[width=2.9cm]{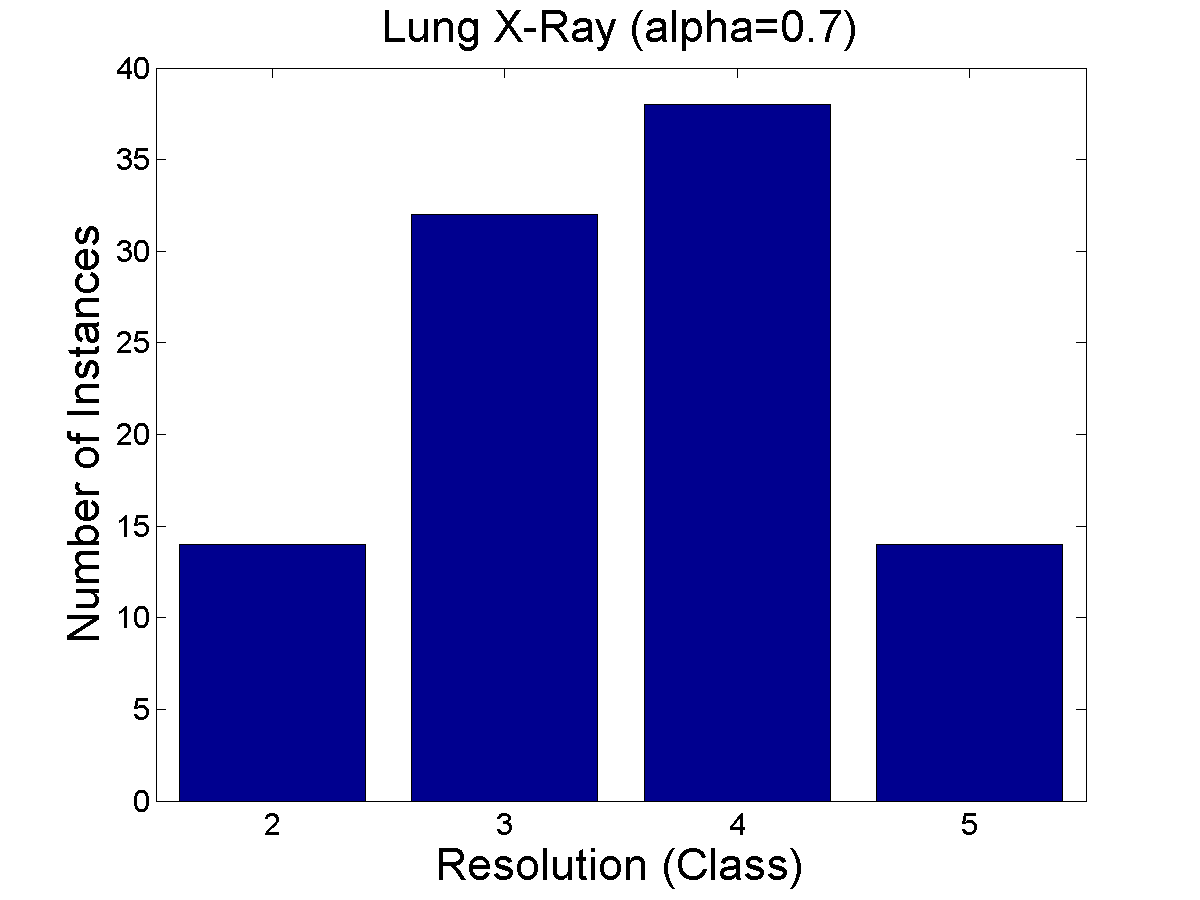} & 
\includegraphics[width=2.9cm]{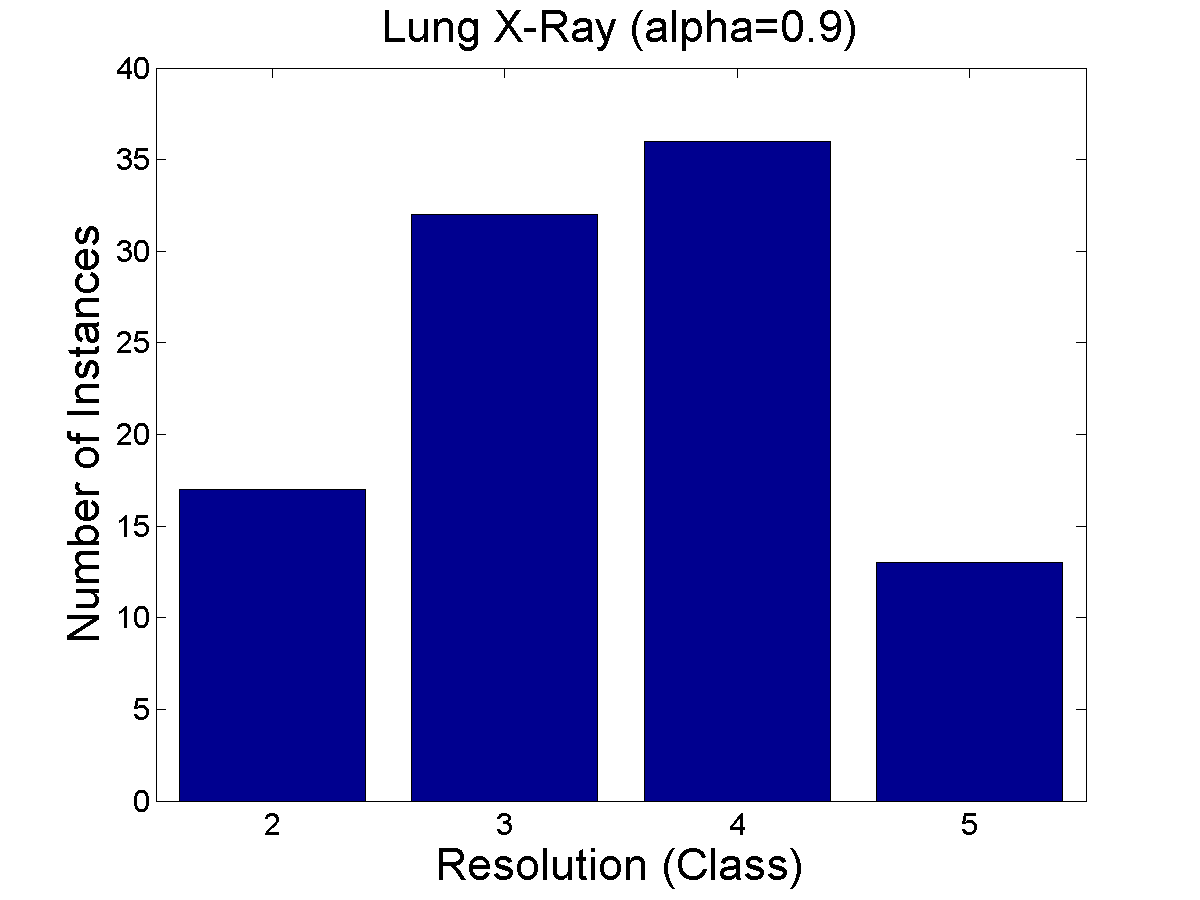}\\
\end{tabular}
\caption{Selected resolution (class) distribution for different values of alpha. First row: breast ultrasound dataset, second row: Liberty dataset, last row: lung X-ray dataset. From left to right: $\alpha$ = 0.1, 0.3, 0.5, 0.7, 0.9.}
\label{fig:Histograms}
\end{center}
\end{figure*}

After each image was segmented at all resolutions, and the segmentation accuracies (dice) and processing times in seconds were recorded, $\omega$ was calculated as described in Section \ref{sec:OptResforTrain}. This subsection presents the results of the assessment of the performance, with respect to segmentation accuracy and speed, of the trade-off measure with different choices of $\alpha$. These values were compared against the accuracies obtained when the images were segmented at the original resolution (level 0), the minimum resolution (level 5), and the \textbf{peak resolutions}, which are the resolutions with maximum frequency for being the best resolution for a given dataset (i.e., the mode of a normal or quasi-normal distribution of the best resolutions, see Figure \ref{fig:Histograms}). The purpose of the comparison with the original resolution is to compare the difference between the outcomes of using the framework and not using it. Comparing with the minimum resolution answers the question of the applicability of just selecting the lowest possible resolution. The importance of learning can be observed by comparing the results of the framework with the peak resolutions, as they statistically constitute the best resolution in a given dataset (one may think just using them instead of learning the best one for each image separately would be more reasonable).

Tables \ref{tbl:OptimalAccTimeBU}-\ref{tbl:OptimalAccTimeLungXray} present the results of breast ultrasound, Liberty, and lung X-ray datasets, respectively. The results show that accuracy increases as $\alpha$ increases. The change in the accuracy level with respect to $\alpha$ can be large, as with the Liberty images, or only minimal, as with the lung images. For all datasets, using higher $\alpha$ values obtained better accuracy results than the original resolution and much better accuracy than the coarsest one. The results listed in the tables confirm that segmentation at resolutions other than the original could enhance the results.


\begin{table}
\tiny
\caption{Accuracy levels and processing times at the selected, peak, original, and minimum resolutions for the ``Breast Ultrasound Dataset''}
\label{tbl:OptimalAccTimeBU}
\begin{center}
\resizebox{0.7\columnwidth}{!}{
\begin{tabular}{c|c|c||c|c}
\hline
 & \multicolumn{4}{c}{Breast Ultrasound Dataset} \\
\hline
$\alpha$ & Dice at Selected & Dice at Peak & Time at Selected & Time at Peak \\
\hline
0.1 & 69.88 $\pm$ 27.88 & 65.09 $\pm$ 30.01 & 0.32 $\pm$ 0.34 & 0.82 $\pm$ 0.50 \\
0.3 & 70.73 $\pm$ 27.55 & 65.09 $\pm$ 30.01 & 0.37 $\pm$ 0.41 & 0.82 $\pm$ 0.50 \\
0.5 & 71.14 $\pm$ 27.46 & 65.09 $\pm$ 30.01 & 0.41 $\pm$ 0.42 & 0.82 $\pm$ 0.50 \\
0.7 & 71.65 $\pm$ 27.38 & 65.09 $\pm$ 30.01 & 0.52 $\pm$ 0.58 & 0.82 $\pm$ 0.50 \\
0.9 & 71.93 $\pm$ 27.45 & 65.09 $\pm$ 30.01 & 0.74 $\pm$ 0.99 & 0.82 $\pm$ 0.50 \\
\hline \hline
Original & \multicolumn{2}{c||}{63.52 $\pm$ 28.92}  & \multicolumn{2}{c}{11.52 $\pm$ 7.90}  \\
Minimum  & \multicolumn{2}{c||}{36.65 $\pm$ 33.03}  & \multicolumn{2}{c}{0.20 $\pm$ 0.09}  \\
\hline
\end{tabular}
}
\end{center}
\end{table}

\begin{table}
\tiny
\caption{Accuracy levels and processing times at the selected, peak, original, and minimum resolutions for the ``Liberty Dataset''}
\label{tbl:OptimalAccTimeLiberty}
\begin{center}
\resizebox{0.7\columnwidth}{!}{
\begin{tabular}{c|c|c||c|c}
\hline
 & \multicolumn{4}{c}{Liberty Dataset} \\
\hline
$\alpha$ & Dice at Selected & Dice at Peak & Time at Selected & Time at Peak \\
\hline
0.1 & 72.07 $\pm$ 17.92 & 69.68 $\pm$ 19.74 & 0.33 $\pm$ 0.11 & 0.48 $\pm$ 0.18 \\
0.3 & 75.80 $\pm$ 18.16 & 69.68 $\pm$ 19.74 & 0.48 $\pm$ 0.31 & 0.48 $\pm$ 0.18 \\
0.5 & 76.58 $\pm$ 18.45 & 73.68 $\pm$ 20.37 & 0.57 $\pm$ 0.38 & 1.08 $\pm$ 0.33 \\
0.7 & 77.58 $\pm$ 18.69 & 74.52 $\pm$ 20.92 & 0.84 $\pm$ 0.66 & 3.39 $\pm$ 1.20 \\
0.9 & 78.04 $\pm$ 18.71 & 74.52 $\pm$ 20.92 & 1.05 $\pm$ 0.86 & 3.39 $\pm$ 1.20 \\
\hline \hline
Original & \multicolumn{2}{c||}{74.24 $\pm$ 18.61}  & \multicolumn{2}{c}{13.54 $\pm$ 5.58}  \\
Minimum  & \multicolumn{2}{c||}{62.44 $\pm$ 20.25}  & \multicolumn{2}{c}{0.27 $\pm$ 0.11}  \\
\hline
\end{tabular}
}
\end{center}
\end{table}

\begin{table}
\tiny
\caption{Accuracy levels and processing times at the selected, peak, original, and minimum resolutions for the ``Lung X-Ray Dataset''}
\label{tbl:OptimalAccTimeLungXray}
\begin{center}
\resizebox{0.7\columnwidth}{!}{
\begin{tabular}{c|c|c||c|c}
\hline
 & \multicolumn{4}{c}{Lung X-Ray Dataset} \\
\hline
$\alpha$ & Dice at Selected & Dice at Peak & Time at Selected & Time at Peak \\
\hline
0.1 & 85.41 $\pm$ 5.75 & 84.64 $\pm$ 6.22 & 0.61 $\pm$ 0.09 & 1.21 $\pm$ 0.18 \\
0.3 & 85.52 $\pm$ 5.73 & 84.64 $\pm$ 6.22 & 0.89 $\pm$ 0.43 & 1.21 $\pm$ 0.18 \\
0.5 & 85.58 $\pm$ 5.68 & 84.64 $\pm$ 6.22 & 1.11 $\pm$ 0.99 & 1.21 $\pm$ 0.18 \\
0.7 & 85.62 $\pm$ 5.65 & 84.64 $\pm$ 6.22 & 1.44 $\pm$ 1.48 & 1.21 $\pm$ 0.18 \\
0.9 & 85.62 $\pm$ 5.65 & 84.64 $\pm$ 6.22 & 1.59 $\pm$ 1.67 & 1.21 $\pm$ 0.18 \\
\hline \hline
Original & \multicolumn{2}{c||}{82.42 $\pm$ 5.20}  & \multicolumn{2}{c}{460.84 $\pm$ 63.07}  \\
Minimum  & \multicolumn{2}{c||}{80.19 $\pm$ 8.27}  & \multicolumn{2}{c}{0.61 $\pm$ 0.09}  \\
\hline
\end{tabular}
}
\end{center}
\end{table}

With respect to the running times, it can be seen from the tables that selecting lower values of $\alpha$ results in faster execution times. Compared with the speed at the original resolution, all values of $\alpha$ for the three datasets produce much shorter processing times. For example, for the times resulting from $\alpha=0.9$, which is considered to be the slowest but most accurate, the speed compared with the original resolution increases by 15, 12, and 290 times for the breast ultrasound, Liberty, and lung X-ray, respectively. The acceleration is especially obvious in the case of the lung X-ray dataset, which has large images and was segmented with the inherently slow level set algorithm.

Comparing the selected resolutions with the peak resolutions accuracies at the selected resolutions with all $\alpha$ values are better than those at the peak ones for the three datasets. With respect to speed, except for $\alpha=0.3$ for the Liberty dataset and $\alpha=0.7$ and 0.9  for the lung X-ray dataset, the speed at selected resolutions is faster than that at the peak resolutions. The difference in speed can be up to 2.6 times for the breast ultrasound dataset, and 4 times for the Liberty dataset.

For breast ultrasound, Liberty, and lung X-ray datasets, Figure \ref{fig:Histograms} illustrates the distributions of the resolutions selected based on different $\alpha$ values for the trade-off measure. It can be observed that the distribution varies significantly from dataset to dataset and for different $\alpha$ values within the same dataset. As can also be seen in many of the figures, the class distribution is imbalanced, which creates difficulties with respect to the learning process as discussed previously. As well, no single selected resolution is revealed, even for images from the same dataset, and in no dataset is the original resolution selected for any value of $\alpha$.

The previous results show that selecting resolutions for images in a specific dataset can obtain higher accuracies and faster speeds than fixing one resolution for all images. Segmenting at the original resolution results in lower accuracies and much slower speeds, while segmenting at the lowest resolution will result in much lower accuracies. Even compared with the peak resolutions, resolutions selected based on the trade-off measure obtained better results.

Note that for the lung X-ray dataset, the difference in accuracies among the selected, peak, original, and minimum resolutions is very small. Moreover, the difference in speed between selected and peak resolution is also small, and the peak resolution is faster than the selected ones for $\alpha$=0.7 and 0.9. The reason for this is the extreme similarity of the objects (the lungs) among all images in the dataset. This suggests that learning to select best resolutions for image segmentation might not be suitable for such cases.

\subsection{Classifier Performance}
This section presents an evaluation of the performance of RAMOBoost for resolution selection. As explained in Section \ref{sec:featureExtraction}, Local Binary Patterns (LBP) features were used for the learning process. The results obtained were compared with those produced by AdaBoost, SVM, and SVM with training data re-balanced via ADASYN (SVM-ADASYN). The targets (labels) were defined based on the trade-off measure $\omega$ (Eq. \ref{eq:metric}). In these experiments, five values of $\alpha$ were used: 0.1, 0.3, 0.5, 0.7, and 0.9. $F_1$\emph{-measure} (Eq. \ref{eq:FMeasure}) was used to evaluate the performance per class. The numerical results are listed in Tables \ref{tbl:ClassPerfBreastUlt}, \ref{tbl:ClassPerfLiberty}, and \ref{tbl:ClassPerfLungXray} in the Appendix. Figures \ref{fig:ClassPerfBU_LBP_CM}-\ref{fig:ClassPerfLung_LBP_CM} illustrate the performance for $\alpha$=0.7 along the confusion matrices. Because the presented classifiers' performances are the average of 10 runs, the mean and standard deviation of 10 confusion matrices are calculated. The values were rounded to the nearest integer for easier interpretation. The number of instances is different for each class. Figure \ref{fig:Histograms} show the number of instances for each class for the three datasets.

For the three datasets, Figure \ref{fig:Histograms} reveals an imbalance in class distribution. Two of the methods used are designed specifically for imbalanced classification problems: RAMOBoost and ADASYN. The use of these methods was assessed with respect to the effect on the classification of minority class examples. The detailed numerical results are presented in the Appendix.

\textbf{Breast Ultrasound Dataset} - Figure \ref{fig:ClassPerfBU_LBP_CM} shows a comparison of the values of the $F_1$\emph{-measure} obtained from the four learning methods for the breast ultrasound dataset with $\alpha$=0.7. It can be observed that RAMOBoost obtained the best overall results, closely followed  by AdaBoost. The boosting algorithms significantly outperformed both SVM and SVM-ADASYN. The results reveal that SVM-ADASYN is slightly better than SVM with respect to classifying minority class samples, as in class 2 with $\alpha$=0.3 and classes 1 and 5 with $\alpha$=0.9. RAMOBoost is superior to AdaBoost in classifying minority classes, especially for class 5 with $\alpha$=0.5, classes 1 and 5 with $\alpha$=0.7 and class 1 with $\alpha$=0.9.

\begin{figure}[htb]
\begin{center}
\resizebox{\columnwidth}{!}{
\begin{tabular}{cc}
\subfloat{\includegraphics[width=0.4\textwidth]{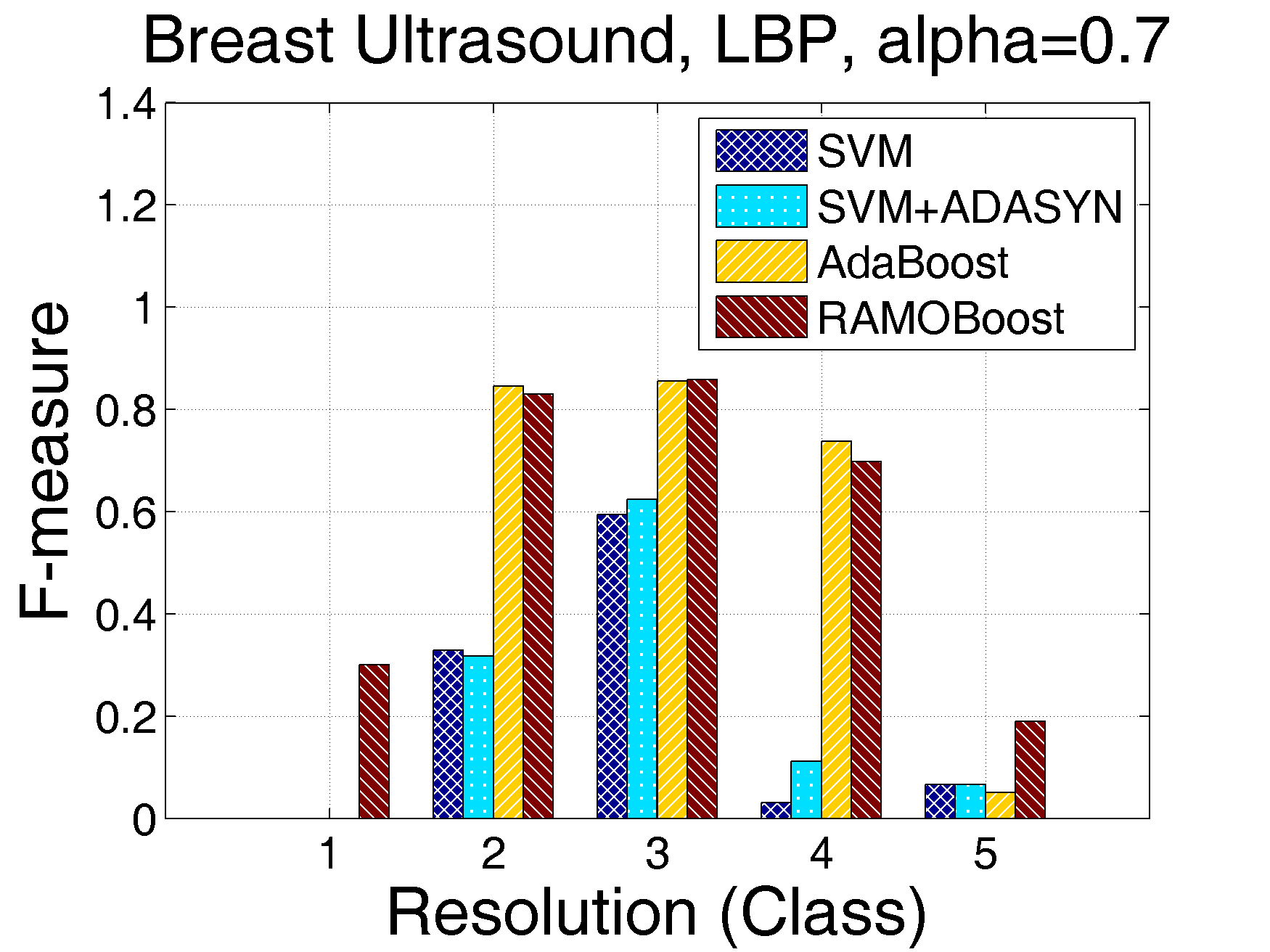}} &
\subfloat{\includegraphics[width=0.3\textwidth]{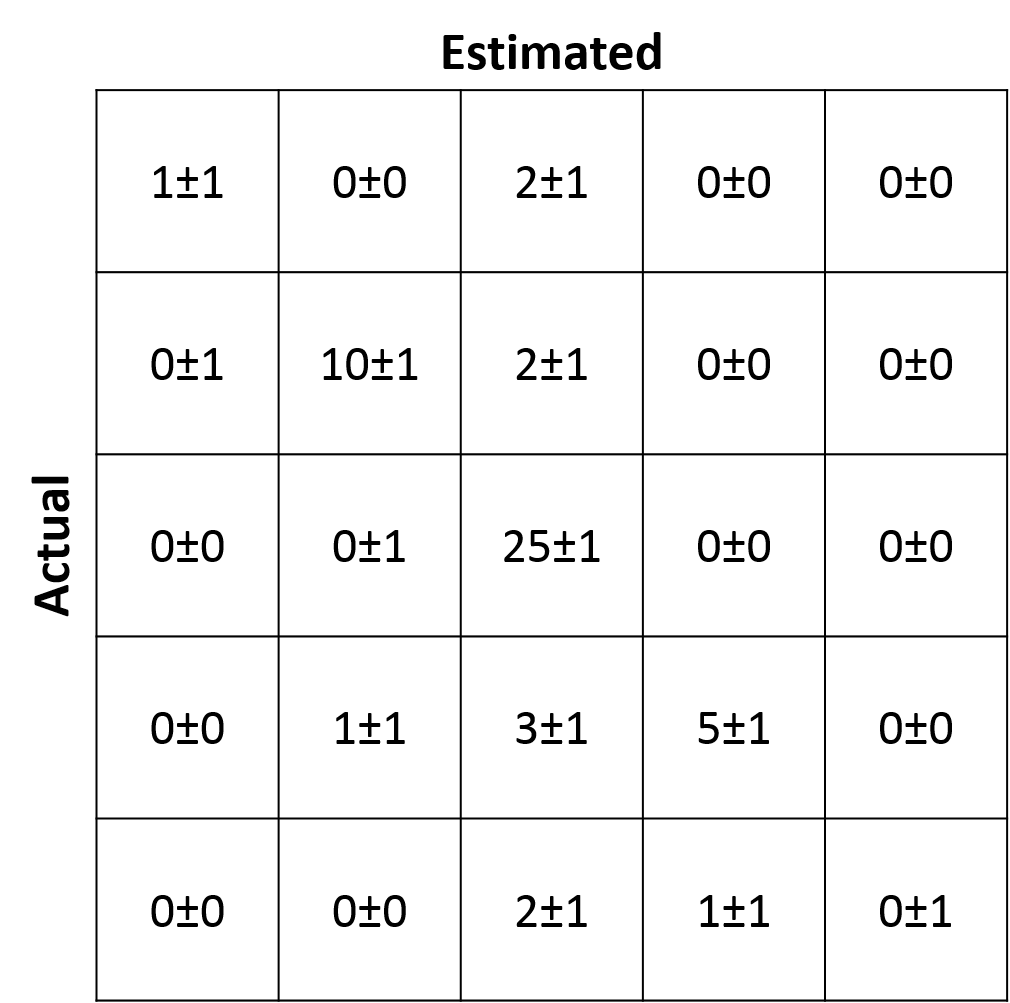}} \\
\end{tabular}
}
\caption{Left: performance of different classifiers with $\alpha$=0.7 for the breast ultrasound dataset: LBP features; right: confusion matrix of RAMOBoost.}
\label{fig:ClassPerfBU_LBP_CM}
\end{center}
\end{figure}

\textbf{Liberty Dataset} - The classification performance for the Liberty dataset is presented in Figure \ref{fig:ClassPerfLiberty_LBP_CM} for $\alpha$=0.7. RAMOBoost provided the best overall performance, with AdaBoost producing slightly worse results. Both boosting methods are far superior to SVM and SVM-ADASYN. For example, class 4 with $\alpha$=0.7 could not be correctly classified by either SVM or SVM-ADASYN, while it was accurately classified by both boosting algorithms. SVM and SVM-ADASYN offer similar performance levels. It should be noted how well the boosting algorithms perform with respect to the imbalanced class distributions, as with $\alpha$=0.7 and $\alpha$=0.9.

\begin{figure}[htb]
\begin{center}
\resizebox{\columnwidth}{!}{
\begin{tabular}{cc}
\subfloat{\includegraphics[width=0.4\textwidth]{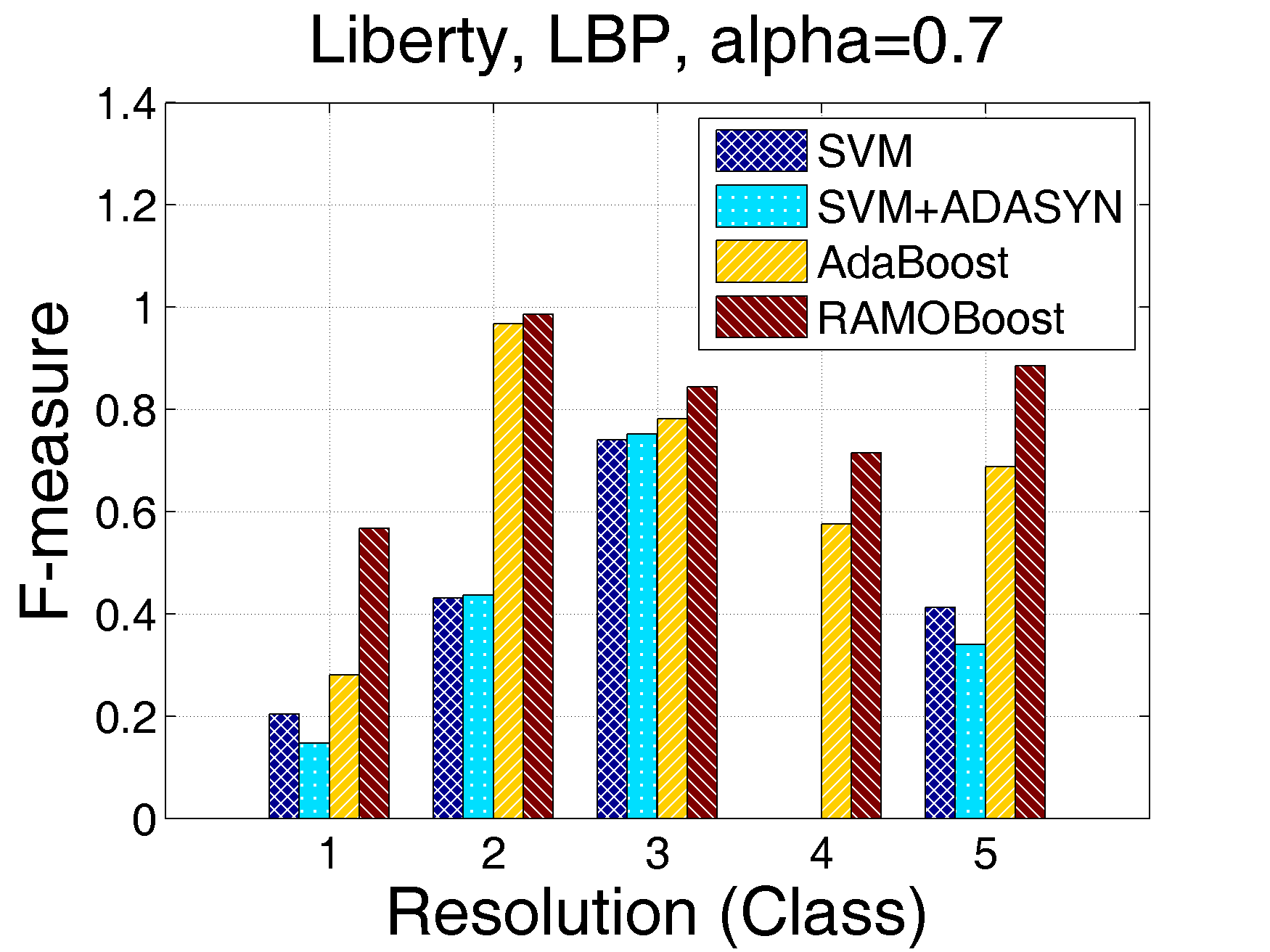}} &
\subfloat{\includegraphics[width=0.3\textwidth]{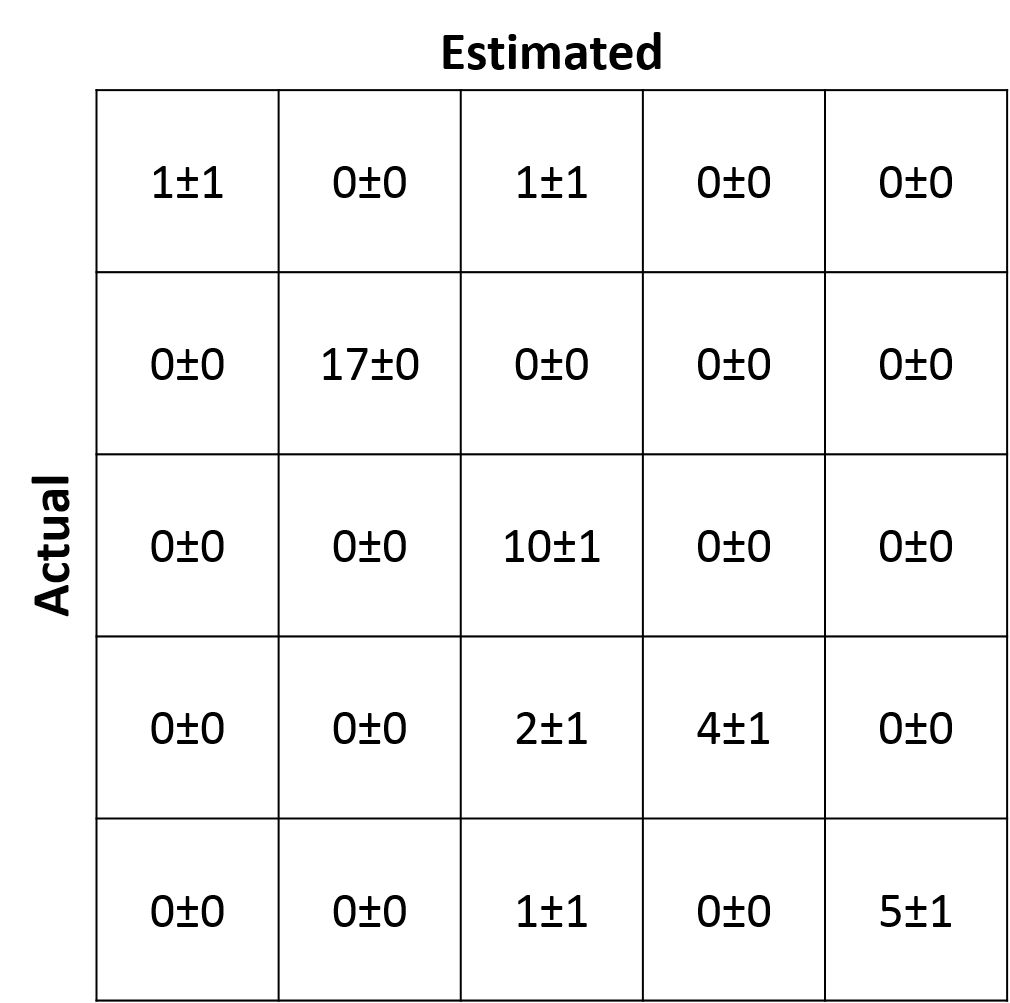}} \\
\end{tabular}
}
\caption{Left: performance of different classifiers with $\alpha$=0.7 for the Liberty dataset: LBP features; right: confusion matrix of RAMOBoost.}
\label{fig:ClassPerfLiberty_LBP_CM}
\end{center}
\end{figure}

\textbf{Lung X-ray Dataset} - A comparison of the performance of the different classification algorithms for the lung X-ray dataset is shown in Figure \ref{fig:ClassPerfLung_LBP_CM} for $\alpha$=0.7. As with the previous two datasets, RAMOBoost provides the best classification accuracy, followed closely by AdaBoost. SVM and SVM-ADASYN generally produce similar accuracy levels. It can be observed how SVM-ADASYN led to better results than SVM for minority classes such as class 5 with $\alpha$=0.1 and class 2 with $\alpha$=0.5. As with the other datasets, RAMOBoost and AdaBoost result in far superior performance for minority classes estimation. It should be noted that class 2 with $\alpha$=0.3 has only one instance, so it is impossible for it to be learned with any method.
\begin{figure}[htb]
\begin{center}
\resizebox{\columnwidth}{!}{
\begin{tabular}{cc}
\subfloat{\includegraphics[width=0.4\textwidth]{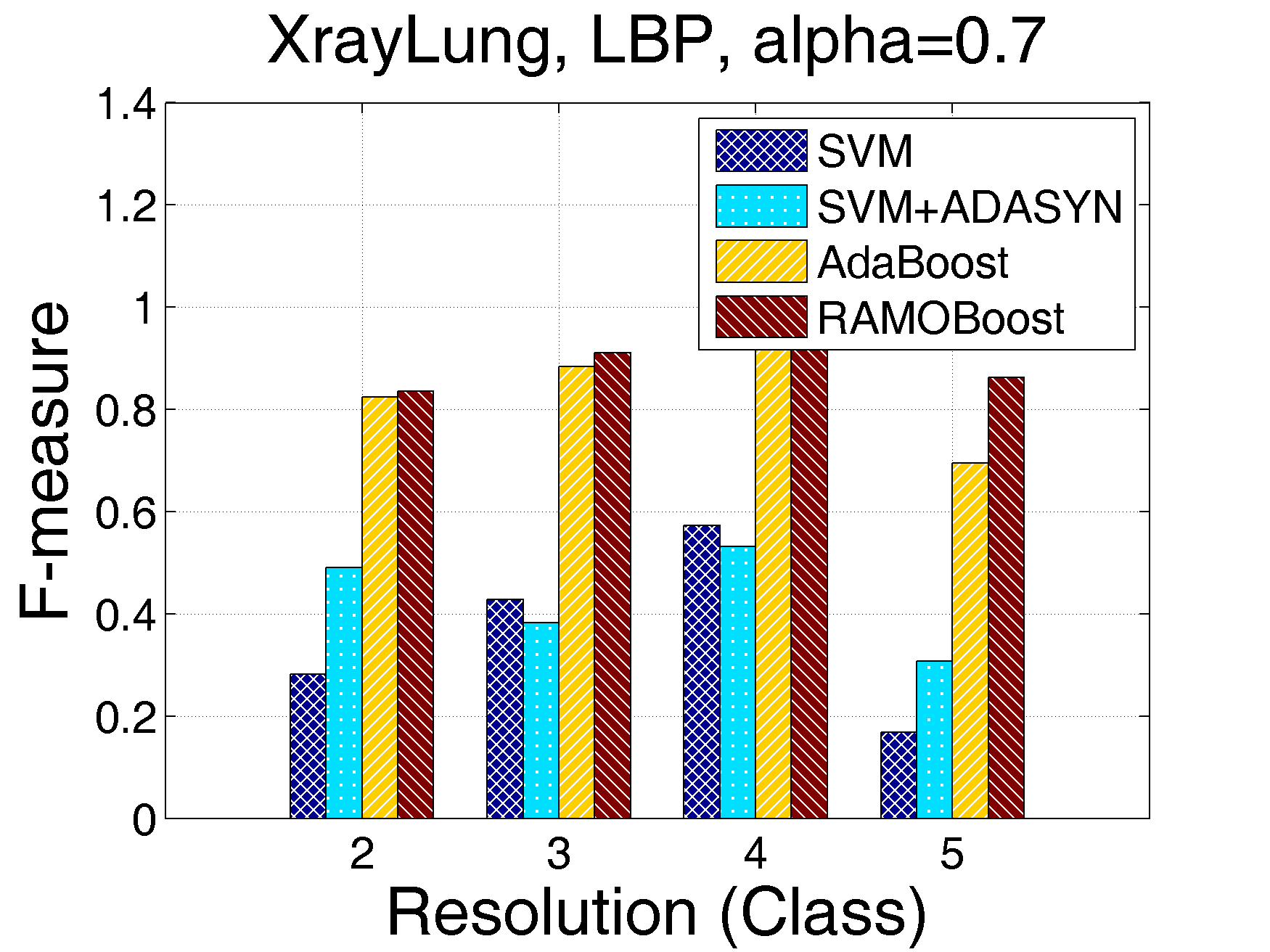}} &
\subfloat{\includegraphics[width=0.3\textwidth]{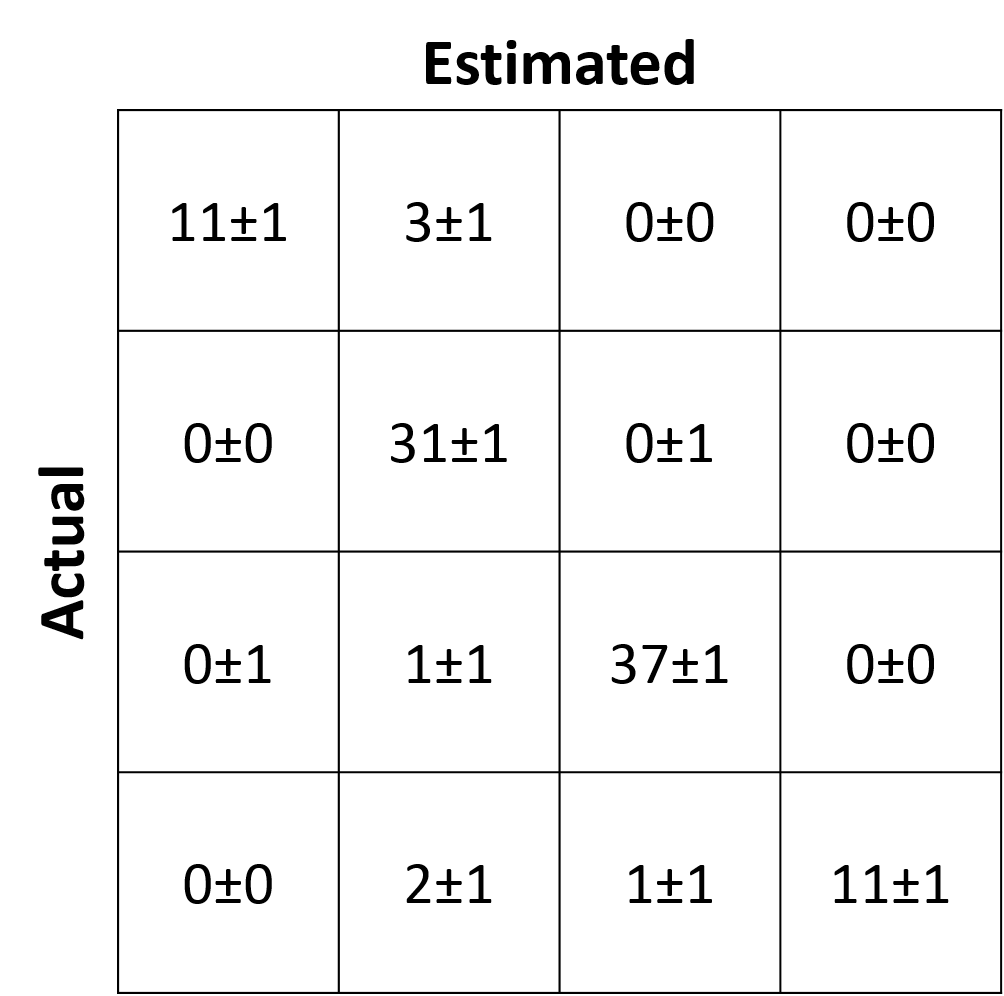}} \\
\end{tabular}
}
\caption{Left: performance of different classifiers with $\alpha$=0.7 for the lung X-ray dataset: LBP features; right: confusion matrix of RAMOBoost.}
\label{fig:ClassPerfLung_LBP_CM}
\end{center}
\end{figure}

\textbf{Observations} - The boosting algorithms greatly outperformed both SVM algorithms. As well, RAMOBoost has better classification performance than AdaBoost, especially for minority classes. This is due to the adaptability of RAMOBoost with respect to imbalanced class distribution.

SVM-ADASYN performs better than SVM for minority classes but slightly worse for majority ones. This shows that using ADASYN to re-balance class distribution may improve the classification of minority classes, but at the cost of affecting the performance for majority ones.

It can be noted from the confusion matrices that the classification performance is good, and the misclassified instances are usually classified as a resolution next to the actual ones, which decreases the impact of misclassification on accuracy and time as shown in the next section. Furthermore, the low standard deviations in the confusion matrices indicate that the classification results are robust and consistent.

\subsection{Impact of Misclassification}
An important factor to assess is the impact of resolution misclassification on segmentation accuracy and speed. The estimated resolutions are compared with the original, minimum, and peak resolutions with respect to both accuracy and speed of segmentation. Although the selected resolutions obtained based on the trade-off measure $\omega$ are not available in real-life situations, they have been included here for reference. Paired t-tests with a 95\% confidence level were performed in order to evaluate the statistical significance of the differences. The effect of the outcomes obtained with RAMOBoost were analyzed.

\textbf{Breast Ultrasound Dataset} -- The findings listed in Tables \ref{tbl:ImpactAccurBreastUlt} and \ref{tbl:ImpactTimeBreastUlt} indicate the impact of misclassification on accuracy and time for the breast ultrasound dataset. As can be observed, the accuracy at the estimated resolutions is better than that at the original resolutions for all $\alpha$ values, and is statistically significant for $\alpha$=0.3, 0.5, 0.7, and 0.9. The estimated resolutions with RAMOBoost provide substantially better accuracy levels than those at minimum resolutions, and are better than those at the peak resolutions, with $\alpha$=0.5, 0.7, and 0.9 demonstrating significant differences. With respect to speed, the segmentation at the estimated resolutions is much faster than at the original ones, substantially slower than at the minimum resolutions, and faster than the peak resolutions, with $\alpha$=0.1, 0.3, 0.5, and 0.7 demonstrating significant differences.
\begin{table}
\tiny
\caption[Impact of Misclassification of RAMOBoost on ``Accuracy'' for the Breast Ultrasound Dataset.]{Impact of Misclassification of RAMOBoost on Accuracy for the Breast Ultrasound Dataset; Est: estimated, Orig: original, Min: minimum, and Sel: selected.}
\label{tbl:ImpactAccurBreastUlt}
\begin{center}
\resizebox{0.7\columnwidth}{!}{
\begin{tabular}{c|c|c|c|c}
\hline
$\alpha$ & Est/Orig & Est/Min & Est/Peak & Est/Sel \\
\hline
0.1 & 1.35 & 4.32 & 1.23 & 0.96  \\
0.3 & 1.38 & 4.76 & 1.26 & 0.98  \\
0.5 & 1.40 & 4.96 & 1.28 & 0.99  \\
0.7 & 1.40 & 4.78 & 1.27 & 0.97  \\
0.9 & 1.39 & 4.55 & 1.26 & 0.96  \\
\hline
\end{tabular}
}
\end{center}
\end{table}
\begin{table}
\tiny
\caption[Impact of Misclassification of RAMOBoost on ``Time'' for the Breast Ultrasound Dataset.]{Impact of Misclassification of RAMOBoost on Time for the Breast Ultrasound Dataset; Est: estimated, Orig: original, Min: minimum, and Sel: selected.}
\label{tbl:ImpactTimeBreastUlt}
\begin{center}
\resizebox{0.7\columnwidth}{!}{
\begin{tabular}{c|c|c|c|c}
\hline
$\alpha$ & Orig/Est & Min/Est & Peak/Est & Sel/Est \\
\hline
0.1 & 22.41 & 0.42 & 1.64 & 0.98  \\
0.3 & 21.19 & 0.40 & 1.55 & 0.99  \\
0.5 & 19.97 & 0.37 & 1.45 & 0.97  \\
0.7 & 18.47 & 0.34 & 1.34 & 1.07  \\
0.9 & 15.47 & 0.29 & 1.12 &  0.97 \\
\hline
\end{tabular}
}
\end{center}
\end{table}

\textbf{Liberty Dataset} -- The impact of misclassification on accuracy and time for the Liberty dataset is shown in Tables \ref{tbl:ImpactAccurLiberty} and \ref{tbl:ImpactTimeLiberty}. It can be seen that the accuracy at the estimated resolutions is almost identical to the accuracy at the original ones, with the estimated resolutions only slightly worse than the original ones at $\alpha$=0.1, and slightly better at the remaining values of the $\alpha$ settings. The accuracy at the estimated resolutions is significantly higher than at the minimum ones and better than the peak resolutions. The speed at estimated resolutions is much faster than at the original ones, but significantly slower than at the minimum resolutions. For $\alpha$=0.1 and 0.3, the segmentation at the estimated resolutions is slower than the ones at the peak resolutions, but it is much faster for $\alpha$=0.5, 0.7, and 0.9.
\begin{table}
\tiny
\caption[Impact of Misclassification of RAMOBoost on ``Accuracy'' for the Liberty Dataset.]{Impact of Misclassification of RAMOBoost on Accuracy for the Liberty Dataset; Est: estimated, Orig: original, Min: minimum, and Sel: selected.}
\label{tbl:ImpactAccurLiberty}
\begin{center}
\resizebox{0.7\columnwidth}{!}{
\begin{tabular}{c|c|c|c|c}
\hline
$\alpha$ & Est/Orig & Est/Min & Est/Peak & Est/Sel \\
\hline
0.1 & 0.99 & 1.22 & 1.05 & 1.00 \\
0.3 & 1.02 & 1.26 & 1.09 & 0.98 \\
0.5 & 1.03 & 1.29 & 1.05 & 0.99 \\
0.7 & 1.04 & 1.29 & 1.04 & 0.98 \\
0.9 & 1.04 & 1.29 & 1.04 & 0.98 \\
\hline
\end{tabular}
}
\end{center}
\end{table}
\begin{table}
\tiny
\caption[Impact of Misclassification of RAMOBoost on ``Time'' for the Liberty Dataset.]{Impact of Misclassification of RAMOBoost on Time for the Liberty Dataset; Est: estimated, Orig: original, Min: minimum, and Sel: selected.}
\label{tbl:ImpactTimeLiberty}
\begin{center}
\resizebox{0.7\columnwidth}{!}{
\begin{tabular}{c|c|c|c|c}
\hline
$\alpha$ & Orig/Est & Min/Est & Peak/Est & Sel/Est \\
\hline
0.1 & 23.40 & 0.45 & 0.83 & 0.99 \\
0.3 & 19.44 & 0.39 & 0.70 & 1.02 \\
0.5 & 17.38 & 0.36 & 1.42 & 0.99 \\
0.7 & 14.33 & 0.31 & 3.66 & 1.03 \\
0.9 & 12.00 & 0.25 & 3.08 & 0.98 \\
\hline
\end{tabular}
}
\end{center}
\end{table}

\textbf{Lung X-ray Dataset} -- Tables \ref{tbl:ImpactAccurLungXray} and \ref{tbl:ImpactTimeLungXray} present the impact of misclassification on accuracy and time for the lung X-ray dataset. The accuracy at the estimated resolutions is statistically better than at the original, minimum, and peak resolutions. The running time is substantially faster at the estimated resolutions than at the original resolutions and slower than at the minimum and peak resolutions.
\begin{table}
\tiny
\caption[Impact of Misclassification of RAMOBoost on ``Accuracy'' for the Lung X-Ray Dataset.]{Impact of Misclassification of RAMOBoost on Accuracy for the Lung X-Ray Dataset; Est: estimated, Orig: original, Min: minimum, and Sel: selected.}
\label{tbl:ImpactAccurLungXray}
\begin{center}
\resizebox{0.7\columnwidth}{!}{
\begin{tabular}{c|c|c|c|c}
\hline
$\alpha$ & Est/Orig & Est/Min & Est/Peak & Est/Sel \\
\hline
0.1 & 1.04 & 1.07 & 1.01 & 1.00 \\
0.3 & 1.04 & 1.07 & 1.01 & 1.00 \\
0.5 & 1.04 & 1.07 & 1.01 & 1.00 \\
0.7 & 1.04 & 1.08 & 1.01 & 1.00 \\
0.9 & 1.04 & 1.07 & 1.01 & 1.00 \\
\hline
\end{tabular}
}
\end{center}
\end{table}
\begin{table}
\tiny
\caption[Impact of Misclassification of RAMOBoost on ``Time'' for the Lung X-Ray Dataset.]{Impact of Misclassification of RAMOBoost on Time for the Lung X-Ray Dataset; Est: estimated, Orig: original, Min: minimum, and Sel: selected.}
\label{tbl:ImpactTimeLungXray}
\begin{center}
\resizebox{0.7\columnwidth}{!}{
\begin{tabular}{c|c|c|c|c}
\hline
$\alpha$ & Orig/Est & Min/Est & Peak/Est & Sel/Est \\
\hline
0.1 & 298.86 & 0.40 & 0.78 & 1.00 \\
0.3 & 282.12 & 0.37 & 0.74 & 1.01 \\
0.5 & 265.23 & 0.35 & 0.70 & 1.01 \\
0.7 & 248.50 & 0.33 & 0.65 & 1.01 \\
0.9 & 235.63 & 0.32 & 0.62 & 0.98 \\
\hline
\end{tabular}
}
\end{center}
\end{table}

\textbf{Observations} - Several trends can be recognized by observing the results:

\begin{itemize}
\item Except at $\alpha$=0.1 for the Liberty dataset, the overall accuracy at the estimated resolutions is always superior to that at the original ones. In some cases, the difference in accuracy can be significant, as in the breast ultrasound and lung X-ray datasets.
\item The accuracy at the estimated resolutions is always significantly higher than at the minimum resolutions.
\item The accuracy at the estimated resolutions is always higher than that at the peak ones.
\item The speed at the estimated resolutions is always much faster than at the original resolutions.
\item The speed at the minimum resolutions is always substantially faster than at the estimated resolutions, but it must be remembered that the extreme speed at the minimum resolutions is accompanied by a considerable (mostly unacceptable) reduction in accuracy.
\item Except with the lung X-ray dataset, the speed at the estimated resolutions is almost always faster than that at the peak ones. Many cases are significantly faster, such as $\alpha$=0.1, 0.3, 0.5, and 0.7 for the breast ultrasound dataset, and $\alpha$=0.5, 0.7, and 0.9 for the Liberty dataset.
\end{itemize}
	
When these observations are considered, it is obvious that even with occasional misclassification, the worst case is that the estimated resolutions produce the same degree of accuracy as the original resolutions but at significantly faster speeds. For two of the datasets, the accuracy at the estimated resolutions is mostly higher than the peak ones, and additionally the speed is faster. This shows that resolution selection for individual images obtains better results than fixing at the peak resolutions. As discussed in Section \ref{sec:propMeasurePerf}, because of the extreme similarity of the images in the lung X-ray dataset, the results in terms of speed compared with the peak resolutions are not satisfactory. This confirms that resolution selection for images of extreme similarity might not be efficient in terms of speed.

\section{Conclusions} \label{sec:Conclusion}
In this work, the problem of resolution selection for image segmentation was highlighted and discussed. We showed that even images from the same dataset could require different  resolutions for a fast and accurate segmentaion. This observation indicates that selecting a fixed resolution for starting coarse-to-fine image segmentation is neither efficient nor effective.

For image segmentation, we identified two main factors that define the best resolution, namely accuracy and speed. Because applications have different objectives, best resolution definitions are accordingly different. We introduced a measure for selecting the best resolution based on the user/application objective. By setting a parameter, one can choose the trade-off between accuracy and time as suitable for the application at hand. Experiments verified that this measure could select resolutions that obtain better accuracy and speed than original resolution.

A framework for automated resolution selection for image segmentation was introduced in this paper. Rank Minority Over-sampling in Boosting (RAMOBoost) was used to learn the best resolutions defined by the trade-off measure. Local Binary Patterns (LBP) features were utilized for the learning. Among four learning methods RAMOBoost obtained the best classification performance, especially for minority classes, closely followed by AdaBoost algorithm.

The misclassification impact on accuracy and time for RAMOBoost was experimentally assessed. The estimated resolutions always lead to much faster segmentation than the original one. In terms of segmentation accuracy, there were no significant difference among estimated and original resolutions in Liberty dataset, while it was substantially better for estimated resolutions in breast ultrasound and lung X-ray datasets.

There is much room for further investigations. Other multiresolution approaches could be investigated. An important criterion would be that the images be subsampled at lower resolutions, excluding methods such as scale-space. Examples of suggested choices are irregular pyramids \cite{IrregularPyramids} and adaptive pyramids \cite{AdaptivePyramids}. An approach involving an adaptive parameter setting per resolution for coarse-to-fine strategy for image segmentation could be studied. It is logical to assume that using the same parameters of a specific image segmentation algorithm at all resolutions would not produce the best results at each resolution: the parameters could be the best for one resolution but be very poor for others. The application of the proposed framework in combination with other image processing applications could be examined as well. For example, it could be applied for image registration, in which case, the trade-off measure could be used as is, because accuracy and time are the two main concerns in this area.

\bibliographystyle{spiejour}   
\bibliography{ref}   

\newpage
\appendix    
\section{Detailed Numerical Results}
This Appendix presents the detailed numerical results of the classifiers performance.

\begin{table}[!h]
\tiny
\caption{Performance of Different Classifiers with Different Values of $\alpha$ for the Breast Ultrasound Dataset (--- denotes irrelevant field)}
\label{tbl:ClassPerfBreastUlt}
\begin{center}
\resizebox{0.7\columnwidth}{!}{
\begin{tabular}{c|c|cccccc}
\hline
 &   & \multicolumn{6}{c}{F-measure}  \\
\hline
$\alpha$ & Learning Alg. & Class 0 & Class 1 & Class 2 & Class 3 & Class 4 & Class 5\\
\hline
\multirow{4}{*}{0.1}                        & SVM        & --- & --- & 0.2760 & 0.6110 & 0.2310 & 0.2430  \\
                                            & SVM+ADASYN & --- & --- & 0.2928 & 0.6145 & 0.2712 & 0.1753  \\
                                            & AdaBoost   & --- & --- & 0.5549 & 0.8445 & 0.8560 & 0.6277  \\
                                            & RAMOBoost  & --- & --- & 0.6488 & 0.8628 & 0.8842 & 0.7530  \\
\hline
\multirow{4}{*}{0.3}                        & SVM        & --- & --- & 0.0000 & 0.6926 & 0.4108 & 0.0000  \\
                                            & SVM+ADASYN & --- & --- & 0.2337 & 0.6565 & 0.3213 & 0.0000  \\
                                            & AdaBoost   & --- & --- & 0.6589 & 0.8666 & 0.8863 & 0.1990  \\
                                            & RAMOBoost  & --- & --- & 0.7718 & 0.8870 & 0.8836 & 0.5067  \\
\hline
\multirow{4}{*}{0.5}                        & SVM        & --- & --- & 0.1896 & 0.6805 & 0.1304 & 0.1000   \\
                                            & SVM+ADASYN & --- & --- & 0.3214 & 0.6230 & 0.2525 & 0.0000   \\
                                            & AdaBoost   & --- & --- & 0.8268 & 0.8797 & 0.8187 & 0.0500   \\
                                            & RAMOBoost  & --- & --- & 0.8809 & 0.9200 & 0.8282 & 0.5000   \\
\hline
\multirow{4}{*}{0.7}                        & SVM        & --- & 0.0000 & 0.3292 & 0.5938 & 0.0308 & 0.0667  \\
                                            & SVM+ADASYN & --- & 0.0000 & 0.3179 & 0.6240 & 0.1115 & 0.0667  \\
                                            & AdaBoost   & --- & 0.0000 & 0.8461 & 0.8550 & 0.7373 & 0.0500  \\
                                            & RAMOBoost  & --- & 0.3000 & 0.8295 & 0.8585 & 0.6984 & 0.1900  \\
\hline
\multirow{4}{*}{0.9}                        & SVM        & --- & 0.0000 & 0.4772 & 0.5317 & 0.2053 & 0.0000  \\
                                            & SVM+ADASYN & --- & 0.0606 & 0.4956 & 0.4488 & 0.2823 & 0.1308  \\
                                            & AdaBoost   & --- & 0.1905 & 0.8923 & 0.8182 & 0.6097 & 0.0000  \\
                                            & RAMOBoost  & --- & 0.6749 & 0.8622 & 0.8390 & 0.5209 & 0.0667  \\
\hline
\end{tabular}
}
\end{center}
\end{table}

\begin{table}[!h]
\tiny
\caption{Performance of Different Classifiers with Different Values of $\alpha$ for the Liberty Dataset (--- denotes irrelevant field)}
\label{tbl:ClassPerfLiberty}
\begin{center}
\resizebox{0.7\columnwidth}{!}{
\begin{tabular}{c|c|cccccc}
\hline
 &   & \multicolumn{6}{c}{F-measure}  \\
\hline
$\alpha$  & Learning Alg. & Class 0 & Class 1 & Class 2 & Class 3 & Class 4 & Class 5  \\
\hline
\multirow{4}{*}{0.1}                        & SVM        & --- & --- & --- & 0.7020 & 0.5797 & 0.6213  \\
                                            & SVM+ADASYN & --- & --- & --- & 0.6899 & 0.6428 & 0.7117  \\
                                            & AdaBoost   & --- & --- & --- & 0.9175 & 0.9643 & 0.9665  \\
                                            & RAMOBoost  & --- & --- & --- & 0.9531 & 0.9707 & 1.0000  \\
\hline
\multirow{4}{*}{0.3}                        & SVM        & --- & --- & 0.3153 & 0.1450 & 0.5342 & 0.1239  \\
                                            & SVM+ADASYN & --- & --- & 0.3460 & 0.1601 & 0.5103 & 0.0903  \\
                                            & AdaBoost   & --- & --- & 0.5790 & 0.8547 & 0.9687 & 0.7325  \\
                                            & RAMOBoost  & --- & --- & 0.5805 & 0.8576 & 0.9744 & 0.9241  \\
\hline
\multirow{4}{*}{0.5}                        & SVM        & --- & --- & 0.6434 & 0.3231 & 0.2479 & 0.0879  \\
                                            & SVM+ADASYN & --- & --- & 0.6793 & 0.3422 & 0.2460 & 0.2237  \\
                                            & AdaBoost   & --- & --- & 0.8963 & 0.8589 & 0.9041 & 0.6574  \\
                                            & RAMOBoost  & --- & --- & 0.9489 & 0.9292 & 0.9251 & 0.8774  \\
\hline
\multirow{4}{*}{0.7}                        & SVM        & --- & 0.2043 & 0.4310 & 0.7401 & 0.0000 & 0.4123  \\
                                            & SVM+ADASYN & --- & 0.1472 & 0.4371 & 0.7523 & 0.0000 & 0.3403  \\
                                            & AdaBoost   & --- & 0.2800 & 0.9680 & 0.7812 & 0.5751 & 0.6880  \\
                                            & RAMOBoost  & --- & 0.5667 & 0.9854 & 0.8440 & 0.7151 & 0.8858  \\
\hline
\multirow{4}{*}{0.9}                        & SVM        & --- & 0.2547 & 0.3628 & 0.6872 & 0.0000 & 0.3671  \\
                                            & SVM+ADASYN & --- & 0.2495 & 0.4323 & 0.7564 & 0.0644 & 0.4559  \\
                                            & AdaBoost   & --- & 0.6401 & 0.9950 & 0.7379 & 0.4267 & 0.6871  \\
                                            & RAMOBoost  & --- & 0.7389 & 0.9974 & 0.7880 & 0.4810 & 0.9273  \\
\hline
\end{tabular}
}
\end{center}
\end{table}

\begin{table}[!h]
\tiny
\caption{Performance of Different Classifiers with Different Values of $\alpha$ for the Lung X-Ray Dataset (--- denotes irrelevant field)}
\label{tbl:ClassPerfLungXray}
\begin{center}
\resizebox{0.7\columnwidth}{!}{
\begin{tabular}{c|c|cccccc}
\hline
 &  & \multicolumn{6}{c}{F-measure}  \\
\hline
$\alpha$  & Learning Alg. & Class 0 & Class 1 & Class 2 & Class 3 & Class 4 & Class 5  \\
\hline
\multirow{4}{*}{0.1} 					    & SVM        & --- & --- & --- & 0.2399 & 0.7345 & 0.0000  \\
                                            & SVM+ADASYN & --- & --- & --- & 0.4086 & 0.5703 & 0.2388  \\
                                            & AdaBoost   & --- & --- & --- & 0.8812 & 0.9532 & 0.8266  \\
                                            & RAMOBoost  & --- & --- & --- & 0.9070 & 0.9658 & 0.9314  \\
\hline
\multirow{4}{*}{0.3}                        & SVM        & --- & --- & 0.0000 & 0.4142 & 0.5805 & 0.0000  \\
                                            & SVM+ADASYN & --- & --- & 0.0000 & 0.4149 & 0.5854 & 0.0206  \\
                                            & AdaBoost   & --- & --- & 0.0000 & 0.9191 & 0.9552 & 0.7743  \\
                                            & RAMOBoost  & --- & --- & 0.0000 & 0.9416 & 0.9709 & 0.8733  \\
\hline
\multirow{4}{*}{0.5}                        & SVM        & --- & --- & 0.0000 & 0.5554 & 0.5752 & 0.0250  \\
                                            & SVM+ADASYN & --- & --- & 0.2594 & 0.3922 & 0.5694 & 0.2498  \\
                                            & AdaBoost   & --- & --- & 0.6713 & 0.9236 & 0.9457 & 0.7999  \\
                                            & RAMOBoost  & --- & --- & 0.5708 & 0.9419 & 0.9501 & 0.8634  \\
\hline
\multirow{4}{*}{0.7}                        & SVM        & --- & --- & 0.2819 & 0.4281 & 0.5732 & 0.1682  \\
                                            & SVM+ADASYN & --- & --- & 0.4908 & 0.3824 & 0.5321 & 0.3076  \\
                                            & AdaBoost   & --- & --- & 0.8245 & 0.8845 & 0.9437 & 0.6946  \\
                                            & RAMOBoost  & --- & --- & 0.8363 & 0.9102 & 0.9650 & 0.8625  \\
\hline
\multirow{4}{*}{0.9}                        & SVM        & --- & --- & 0.4603 & 0.5188 & 0.5077 & 0.1453  \\
                                            & SVM+ADASYN & --- & --- & 0.4159 & 0.3847 & 0.4435 & 0.1901  \\
                                            & AdaBoost   & --- & --- & 0.8691 & 0.8905 & 0.9454 & 0.6450  \\
                                            & RAMOBoost  & --- & --- & 0.8716 & 0.9075 & 0.9531 & 0.7737  \\
\hline
\end{tabular}
}
\end{center}
\end{table}





%

\end{spacing}
\end{document}